\documentclass[10pt,twocolumn,letterpaper]{article}

\usepackage{iccv}
\usepackage{times}
\usepackage{epsfig}
\usepackage{graphicx}
\usepackage{amsmath}
\usepackage{amssymb}

\usepackage[pagebackref=true,breaklinks=true,letterpaper=true,colorlinks,bookmarks=false]{hyperref}

\usepackage{subfigure}
\usepackage{wrapfig}
\usepackage{boldline}
\usepackage[x11names]{xcolor}
\iccvfinalcopy % *** Uncomment this line for the final submission

\def\iccvPaperID{6579} % *** Enter the ICCV Paper ID here

\begin{document}

%%%%%%%%% TITLE
\title{Evolving Search Space for Neural Architecture Search}

\author{Yuanzheng Ci\textsuperscript{1}, Chen Lin\textsuperscript{2}, Ming Sun\textsuperscript{3}, Boyu Chen\textsuperscript{1}, Hongwen Zhang\textsuperscript{4}, Wanli Ouyang\textsuperscript{1} \\
{\textsuperscript{1}The University of Sydney,  \textsuperscript{2}University of Oxford,  \textsuperscript{3}SenseTime Research,  \textsuperscript{4}CASIA} \\
\tt\small \{yuanzheng.ci, boyu.chen, wanli.ouyang\}@sydney.edu.au, chen.lin@eng.ox.ac.uk, \\ \tt\small sunming1@sensetime.com, hongwen.zhang@cripac.ia.ac.cn
% \and
}

% \and

\maketitle
% \thispagestyle{empty}

%%%%%%%%% ABSTRACT
\begin{abstract}
       Automation of neural architecture design has been a coveted alternative to human experts. Various search methods have been proposed aiming to find the optimal architecture in the search space. One would expect the search results to improve when the search space grows larger since it would potentially contain more performant candidates. Surprisingly, we observe that enlarging search space is unbeneficial or even detrimental to existing NAS methods such as DARTS, ProxylessNAS, and SPOS. This counterintuitive phenomenon suggests that enabling existing methods to large search space regimes is non-trivial. However, this problem is less discussed in the literature.

        We present a Neural Search-space Evolution (NSE) scheme, the first neural architecture search scheme designed especially for large space neural architecture search problems. 
        The necessity of a well-designed search space with constrained size is a tacit consent in existing methods, and our NSE aims at minimizing such necessity. Specifically, the NSE starts with a search space subset, then evolves the search space by repeating two steps: 1) search an optimized space from the search space subset, 2) refill this subset from a large pool of operations that are not traversed. 
        We further extend the flexibility of obtainable architectures by introducing a learnable multi-branch setting. With the proposed method, 
        we achieve 77.3\% top-1 retrain accuracy on ImageNet with 333M FLOPs, 
       which yielded a state-of-the-art performance among previous auto-generated architectures that do not involve knowledge distillation or weight pruning. When the latency constraint is adopted, our result also performs better than the previous best-performing mobile models with a 77.9\% Top-1 retrain accuracy. Code is available at \href{https://github.com/orashi/NSE_NAS}{https://github.com/orashi/NSE\_NAS}.
\end{abstract}

%%%%%%%%% BODY TEXT
\section{Introduction}

\begin{figure}[t]
    \centering
    \subfigure[Traditional NAS]{
    \label{tradition_nas}
    \centering
    \includegraphics[width=0.8\linewidth]{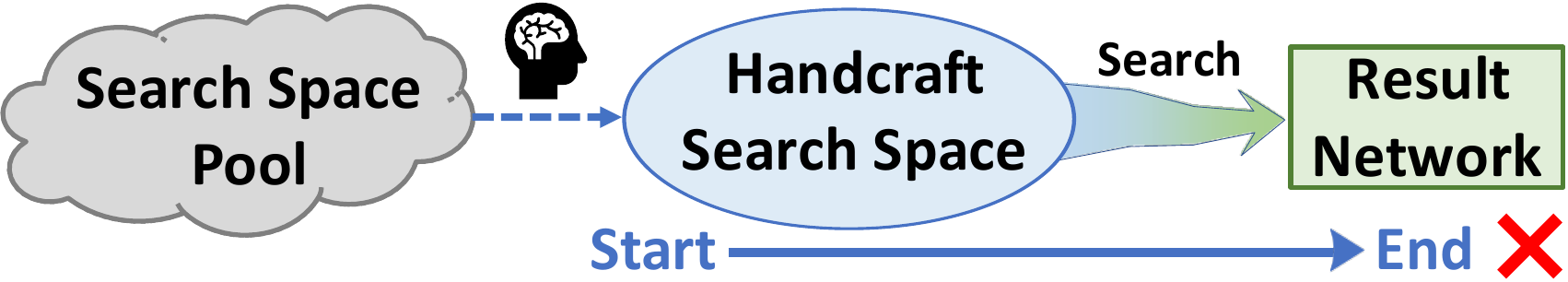}}
    
    \subfigure[Search Space Evolving NAS]{
    \label{space_evo_nas}
    \centering
    \includegraphics[width=0.9\linewidth]{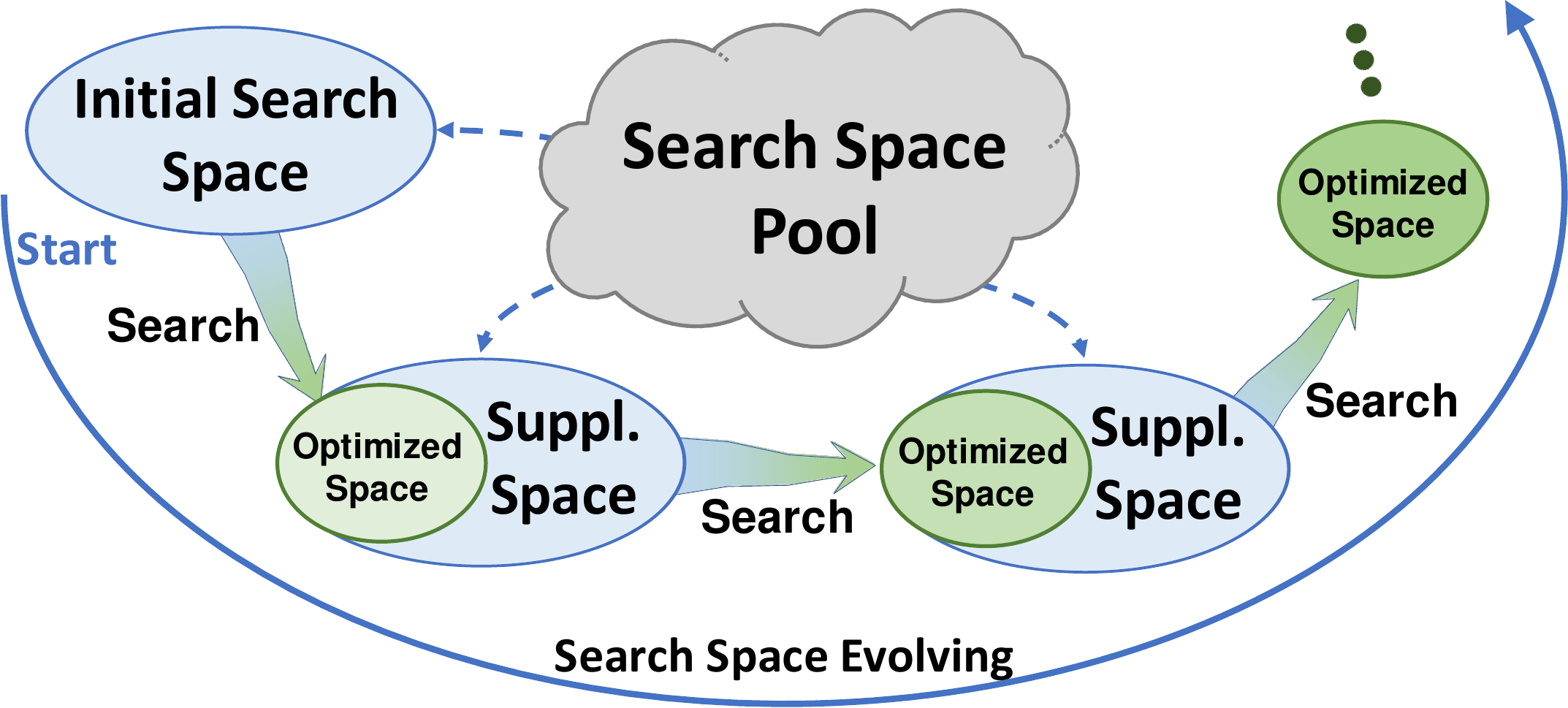}}
    \caption{Comparison of search schemes. (a) Traditional pipeline. (b) Our proposed search space evolving pipeline.}
    % \vspace{-4.mm}
    \label{fig:intro}
\end{figure}

Deep neural networks are prevailing in myriad fields of real-world applications. The emergence of Neural Architecture Search (NAS) has brought up a possibility to automate the customization of deep neural network architectures for specific applications. Researchers have investigated Reinforcement Learning (RL) and Evolutionary Algorithm (EA) based methods~\cite{zoph2016neural,real2018regularized,real2017large,zhong2018practical} to achieve the automation of architecture design.
Weight sharing based methods~\cite{liu2018darts,cai2018proxylessnas,bender2018understanding,chen2021glit,liu2021inception,zhou2020econas,liang2019computation, chen2021bnnas, shen2021towards} that can substantially reduce the computational cost  have  been proposed and became one of the off-the-shelf approaches of NAS research. These methods successfully yielded promising results that have surpassed human-designed architectures~\cite{wistuba2019survey}. 

In addition to the search method, another key component of NAS is the search space.
When compared with early NAS works~\cite{zoph2018learning,real2018regularized}, the quality of the search space has been improved along with the development of search algorithms \cite{radosavovic2019network}. It has been observed that the improvement of search space design imposed a positive effect on the performance of many existing works~\cite{radosavovic2019network, yang2019evaluation}. In particular, the research community has devoted multiple efforts to search space design, from selecting well-suited operations based on the prior knowledge~\cite{tan2019mixconv, guo2019single} to utilizing channel-level fine-grained model search  \cite{wan2020fbnetv2, stamoulis2019single, guo2019single, yu2020bignas, cai2019once, lu2020nsganetv2} over a smaller set of operations. Recent methods often limit the number of candidate operations in each layer to less than ten (excluding decisions on activation functions or SE modules~\cite{hu2018squeeze}). However, such search space improvements fall back into the paradigm of expertise design, which is a countermarch of automated learning of architectures. %\yz{need discussion: automatic principle/automated learning of architecture no need}

A natural question now could be considered is: \emph{can we construct a huge search space which is a super-set of a formerly mentioned space and obtain superior results from it?} If the answer is YES, this approach would solve the problem of search space design by simply using the largest search space one can construct. However, it is less discussed by NAS literature.
%\wl{ (how about: However, the answer is NO, if existing NAS approaches are used.} \yz{a little bit too much?} 
Yu \etal~\cite{yu2019evaluating} and Zhang \etal~\cite{zhang2018you} provide results suggesting that, under the traditional NAS pipeline as shown in Figure~\ref{fig:intro}(a), simply enlarging the search space
%to include more possibilities 
could be detrimental to the final result. 

To look into this issue in detail, we set up a search space consists of 27 distinct operations and then tested 4 reasonably fast NAS algorithms, including DARTS~\cite{liu2018darts}, Proxyless~\cite{cai2018proxylessnas}, SPOS~\cite{guo2019single} and One-Shot~\cite{bender2018understanding}. We show that all of these methods do not hold the
behavior of obtaining better search results with a larger search space. Furthermore, some have their search cost prohibitively high or failed to converge, while others perform poorly even with their training epochs increased (effectiveness of longer training schedule is suggested by~\cite{zhang2020does, bender2020can}). Another relevant technique is search space simplification, which can also assist neural architecture search \cite{zhang2020does, hu2020angle, fang2019betanas}, we will show that such technique is not enough to help NAS algorithm exploit large space effectively.

In this work, we aim at large space neural architecture search by proposing a Neural Search-space Evolution (NSE) scheme.  Instead of directly confronting the negative impacts derived from a large search space, the NSE starts with a search space subset, which is a reasonably sized random subset of the full search space, and search an optimized space from this subset, then refill this subset and repeat the search-then-refill steps as shown in Figure~\ref{fig:intro}(b) to traverse the whole space progressively. 
The resulted NSE enables ever-evolving search space for Neural Architecture Search.

NSE progressively explores extra operation candidates while retaining past knowledge. The search process is constructed as an iterative process to traverse the pending unseen operation candidates. During the iterative process, instead of keeping a single architecture as the intermediate result, we combine all architectures on the Pareto front found by a supernet trained with One-Shot \cite{bender2018understanding} method to obtain an optimized search space, which will be inherited to the next round of search. By maintaining an optimized search space as knowledge, the search process can always proceed with new candidate operations added to the pending list, which means we can always add newly proposed operations in CNN literature that are less verified yet potentially efficient for specific tasks.

To effectively exploit more complex architectures, we further adopt the proposed paradigm to the multi-branch scheme, which has orders of magnitude more distinct structures when compared to its single-branch counterpart. 
Compared with previous single-branch schemes like DARTS that only allow one operation to be selected, the multi-branch scheme allows multiple operations to be selected adaptively. By constructing a probabilistic model for the multi-branch scheme, we can retrieve the fitness of every candidate operation. Operations with fitness lower than a certain threshold will be dropped from the search space so that the complexity of the search space is progressively reduced and the degree of co-adaptation for the rest of the possible path combinations could also be enhanced. 

We conduct experiments on ImageNet \cite{ILSVRC15} with two resource constraints, \ie FLOPs and Latency.
For these two constraints, NAS under our NSE scheme effectively exploited the potential of a very large search space and secured a continual performance increment in the iterative process, leading to state-of-the-art results. 

In summary, the key contributions of this paper are summarized as follows:
\begin{itemize}
    \item We propose NSE, the first neural architecture search scheme that designed especially for large space neural architecture search problems, which empowers NAS to minimize the necessity of dedicated search space design.
    The inheritance property of the evolving process keeps knowledge derived from previous search space while improving that knowledge with new operations added into the current search space. 

    \item We propose a probabilistic modeling of operation-wise fitness for the multi-branch scheme, which makes it feasible to gradually simplify the multi-branch scheme search space as shared weights converge through one-shot training. Such an annealing paradigm gradually simplifies the complexity of the sub-task and helps the remaining
shared weights to be learned better~\cite{zhang2020does}.

\end{itemize}

%------------------------------------------------------------------------

\section{Related Work}
\begin{figure*}[t]
    \centering
    \includegraphics[height=4.6cm]{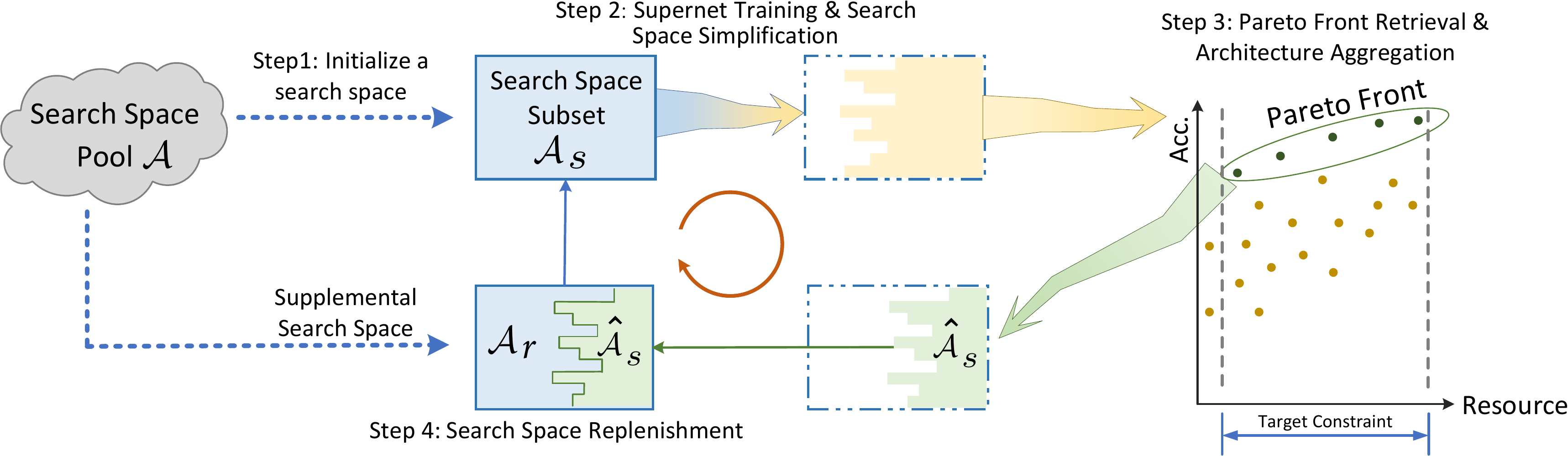}
    \caption{Search space update scheme for proposed approach. A search space subset is retained to incorporate new unseen search space while preserving existing knowledge.}
    % \vspace{-2.0mm}
    \label{fig:search}
\end{figure*}
% \vspace{-15mm}

\subsection{NAS Algorithm Design}
Network architecture search algorithms are essential for finding good architectures. These algorithms are based on Bayesian optimization, Reinforcement Learning (RL), Genetic Algorithm (GA), Weight sharing, and One-shot. There are Bayesian optimization methods that formulate the NAS as a hyperparameter optimization problem and allow to search non-fixed length neural architectures \cite{mendoza2016towards,bergstra2013making}.
RL-based NAS \cite{zoph2016neural,zoph2018learning,zhong2018practical} adopted RL to learn generating the best architecture. GA-based methods \cite{real2017large,real2018regularized} use GA to generate candidate architectures and are also popular with impressive results.
Weight sharing approaches \cite{bender2018understanding,liu2018darts,cai2018proxylessnas,guo2019single,chu2019fairnas,lu2018nsga} utilize a supernet that shares weights among different architectures.  One-shot NAS \cite{bender2018understanding,guo2019single}  by directly training the supernet with dropout, the supernet with shared weights are utilized for predicting model score predictor. However, when new candidate operations are available, these NAS approaches have to restart the search. In contrast, under our NSE scheme, a subset of search space is maintained to efficiently accommodate new candidate operations and inherit the knowledge of previously searched candidate operations. 
%\yz{further discussion for the following? no need} 
Recent works \cite{yu2020bignas, cai2019once, lu2020nsganetv2, mei2019atomnas} also integrated knowledge distillation or weight pruning into the NAS pipeline for instant performance. Nevertheless, these techniques can still be conducted independently with considerable gain \cite{yan2020fp, blalock2020state} and are orthogonal to our focus.
%Recent works \cite{yu2020bignas, cai2019once, lu2020nsganetv2, mei2019atomnas} also integrated knowledge distillation orand weight pruning into the NAS pipeline for instant performance. Nevertheless, \yz{ithese techniques can still be cdonducted separately with considerable gain \cite{yan2020fp, blalock2020state} and areis orthogonal to our focus}.

\subsection{Search Space Design}
The NAS search space is identified to have posed a non-negligible impact on search results \cite{yu2019evaluating,zhang2018you}. Meanwhile, search space design has been improved together with the NAS algorithms \cite{radosavovic2019network}. %\wl{First NAS method}
Zoph \etal~\cite{zoph2016neural} adopted design space with naive building blocks and jump connections. Many approaches repeat the same building block they searched via NAS to construct the network. \cite{zoph2018learning,zhong2018practical,real2018regularized,liu2018darts}. 
For better flexibility, many latter approaches enable searching different operations for different blocks but have to constrain the number of candidate operations for controlling the search space size \cite{tan2018mnasnet,cai2018proxylessnas,wu2018fbnet,guo2019single,fang2019betanas,zhang2019hardware}. % thus the task is transformed into making optimal candidate decisions for every slot in the sequential network backbone,
Orthogonal works that search for fine-grained model adjustments and training configs are also proposed \cite{yu2018slimmable,yu2019universally,mei2019atomnas, wan2020fbnetv2, dai2020fbnetv3}. Recent works \cite{wu2018fbnet,guo2019single,cai2018proxylessnas,tan2018mnasnet,zhang2019hardware,howard2019searching,tan2019mixconv} also include new candidate operations or module designs such as hand-crafted multi-branch cells~\cite{tan2019mixconv}, tree-structure~\cite{cai2018proxylessnas}, shuffle operation \cite{ma2018shufflenet}, Squeeze-and-Excitation (SE) module \cite{hu2018squeeze} and swish activation \cite{ramachandran2017searching}. Our NSE scheme enables neural architecture search with ever-evolving new candidate operations.

%-------------------------------------------------------------------------

\section{Method}

\textbf{Problem Formulation.} 
As the neural architecture commonly uses a feed-forward structure, we represent an overall search space pool $\mathcal{A}$ as a direct acyclic graph (DAG) of $L$ layers, $\bigcup_{l=1}^{L}\mathbf{E}_l$
%\{\mathbf{E}_1, \ldots, \mathbf{E}_L\}
 , where $\mathbf{E}_l$ represents available operations (\eg, $3\times 3$ convolution, pooling, or identity) in the $l$-th layer of DAG. We denote a neural network within the search space as $a=\bigcup_{l=1}^{L}\mathbf{e}_l$
 %\{\mathbf{e}_1, \ldots, \mathbf{e}_L\}
 , where $\mathbf{e}_l \subseteq \mathbf{E}_l $.  
 
 %Let $\mathcal{O} = {op_n}$ for $n=1, \ldots, N$ be the set of $N$ candidate operations (\eg, $3\times 3$ convolution, pooling, or identity), where $n$ denotes the index of the operation. 
 
 %\wl{Define $\mathcal{A}$} 
%\hat{\mathcal{A}_s^{l}} = \bigcup\limits_{p=1}^{P} \{op_{\mathcal{I}_l(n_p)}|s_{\mathcal{I}_l(n_p)}=1\}_{n_p=1}^{K}.

Noticing that multi-branch based networks like Inception~\cite{szegedy2017inception} and ResNeXt~\cite{xie2017aggregated} 
were important inventions to the CNN literature, we include this scheme into our search space evolution. For a network architecture in the \emph{multi-branch} scheme, a layer consists of multiple operations $\{op_{n}\}$ selected from $N$ operation candidates, 
\ie $\mathbf{e} = \mathbf{o}_{\mathbf{g}} = \{op_{n}| g_{n}=1, n\in\{1, \ldots, N\} \}$, 
%\ie $\mathbf{e}_l = \{op_{\mathcal{I}_l(n_l)}| s_{\mathcal{I}_l(n_l)}=1, n_l=1, \ldots, N_l \}$, 
where $\mathbf{g}$ denotes a specific set of operation configuration  $\{g_{n}\}$ and the binary gates $g_{n}\in \{0, 1\}$ denotes whether the $n$-th operation is selected or not. 
%where $\mathcal{I}_l(n_l)$ denotes the index of the $n_l$-th selected operation for the $l$-th layer and the selector $s_{\mathcal{I}_l(n_l)}\in \{0, 1\}$ denotes whether the $n_l$-th operation for the $l$-th layer is selected or not. 
In this case, the number of selected operations within $\mathbf{o}_{\mathbf{g}}$ is $\sum_{n=1}^{N} g_{n}$, and the total amount of possible operation configurations (\ie combinations) is $2^{N}$. 

%Given a specific operation configuration $\{s_{\mathcal{I}_l(n_l)}\}_{n_l=1}^{N_l}$ of the $l$-th layer, 
%we denote this configuration as $\mathbf{s}_{l,l_c}$ and the corresponding selected operations as $\mathbf{e}_{\mathbf{s}_{l,l_c}}$, where $l_c=\sum_{n_l=1}^{N_l}2^{n_l-1}{s_{\mathcal{I}_l(n_l)}}$. 
%   A layer $l$ with configuration $\mathbf{s}_{l,l_c}$ is also denoted as $\mathbf{e}_{\mathbf{s}_{l,l_c}}$.
%Note that the \emph{single-branch} scheme in existing works \cite{bender2018understanding,liu2018darts,cai2018proxylessnas,bender2018understanding,guo2019single} is a spacial case with $N_l=1$ under our formulation.

%The goal of the proposed pipeline is to learn is to find an optimal architecture $a^* \subseteq \mathcal{A}$ that has the highest accuracy over the validation set. \lc{is this still the goal know}

The goal of the proposed pipeline is to explore an extremely large search space by evolving through the search space subsets and looking for the best-performing architecture $a^* \subseteq \mathcal{A}$ at the same time. 
  
 \subsection{Overview}
 \label{sec:overview}

We formulate our search space evolving NAS pipeline as an adaptive process that is capable of exploring from a stream of search space subset replenishment, as depicted in Figure~\ref{fig:search}. The pipeline of NSE scheme is as follows:

Step 1. Initially, a search space subset $\mathcal{A}_s$ is sampled from the overall search space pool $\mathcal{A}$. This is achieved by randomly sampling $K$ candidate operations for each layer. 
In practice, the search space subset $\mathcal{A}_s$ is much smaller than the whole set $\mathcal{A}$ to avoid the large search space dilemma we mentioned. 
It consists of candidate operations currently being considered for NAS algorithms to search; we believe this search space subset, limited in size, is naturally easier for a search algorithm to handle.

Step 2. Supernet training (Section \ref{sec:mb}) and  search space simplification (Section \ref{sec:sss}) are alternately conducted within search space subset $\mathcal{A}_s$. If the fitness of an operation is found to be lower than a certain threshold during the supernet training, then this operation will be dropped from the search space subset.

%Network architectures in multi-branched manner are searched in sub search space $\mathcal{A}_s$, this process consists of supernet training, prunining and sub-network evaluation (Section \ref{sec:Train}). The output of this step is the Pareto front for $\mathcal{A}_s$.

Step 3. The optimized search space subset is obtained by sampling architectures from the search space subset $\mathcal{A}_s$ and evaluating them (Section~\ref{sec:pofront}). Specifically, the fitness of these sampled architectures is evaluated by validation accuracy to retrieve a Pareto front. We aggregate architectures at the Pareto front and obtain the optimized search space subset $\hat{\mathcal{A}_s}$, which is the smallest network search space containing all architectures in the Pareto Front.

Step 4. A new search space $\mathcal{A}_s'$ is constructed (Section~\ref{sec:Inher}).  $\mathcal{A}_s'$ is a combination of optimized search space subset $\hat{\mathcal{A}_s}$ from Step 3 and a supplemental subset of search space $\mathcal{A}_r$.  $\mathcal{A}_r$ is a replenishment sampled from $\mathcal{A}$ to ensure that there are still $K$ candidate operations in each layer. The traversed search space is excluded from $\mathcal{A}_r$ so that the same operation in a layer will not be sampled twice.
In this way, the accumulated knowledge in $\hat{\mathcal{A}_s}$ is inherited. Let $\mathcal{A}_s=\mathcal{A}_s'$ , re-initialize all weights, then go to Step 2. When there is not enough operation remained in $\mathcal{A}$,
the loop ends, and the Pareto front for $\mathcal{A}_s$ in Step 2 is used as the final result of the architecture search.
 %the loop ends when there are no enough operations remain in $\mathcal{A}$ to keep $\mathcal{A}_s'$ have $K$ candidate operations available for each layer

\begin{figure}[t]
\centering
\includegraphics[height=3.8cm]{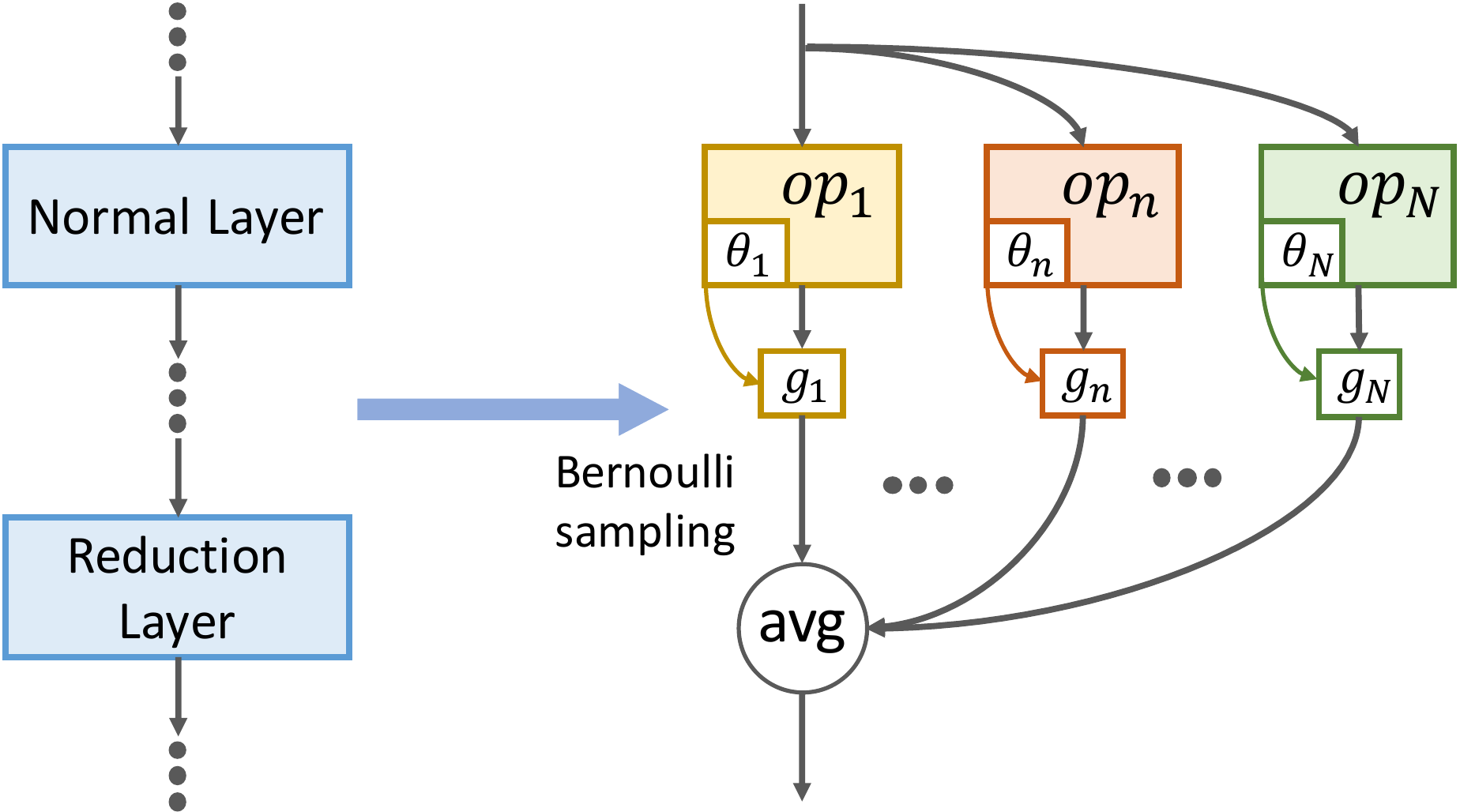}
\caption{Forwarding scheme for multi-branch paths.}
% \vspace{-4mm}
\label{fig:branch}
\end{figure}

\subsection{Supernet Training}
% \label{sec:Train}
\label{sec:mb}
%\subsubsection{Mulit-branch Structure}
%Given a sub search space $\mathcal{A}_s$, we introduce how to train the supernet and prune the search space based on importance indicators in this section, which corresponds to the Step 2.
% Section \ref{sec:overview}. %\wl{Please confirm if this corresponds to only step 2, if not, revise this sentence.}

%\lc{To construct a extremely-sized search space} \lc{why?}, we not only involve more candidate operations into the search space but also 

%\wl{explain how you obtain the size, at least in the supp. I think this can be introduced elsewhere, not necessarily here.}

% apart from simply adding more candidate operations to the search space, we extended the conventional single-branch structure to multi-branch, the search space is thus enlarged by orders of magnitudes as is shown in Table [TODO]. 
\subsubsection{Forward Path in Multi-Branch Scheme} 
Given an input feature map $\mathbf{X}$, the output of a branch $op_n$ can be written as $g_n op_n(\mathbf{X})$.
% denoted as $g_i o_i(x)$. 
As illustrated in Figure~\ref{fig:branch}, the output of a multi-branch layer under the configuration of $\mathbf{g}=\{g_n\}$ can be defined as follows:

%given  and input feature $\mathbf{X}^l$, 

\begin{equation}
\label{equ:output2}
\mathbf{o}_{\mathbf{g}}(\mathbf{X}) = \frac{1}{\sum_{n\in\mathbf{K}} g_{n}}\sum_{n\in\mathbf{K}}g_{n}op_{n}(\mathbf{X}),
\end{equation}
% Since there are $\sum_{n_l=1}^{N_l} s_{\mathcal{I}_l(n_l)}\geq 1$ branches selected for generating the output in Equation (\ref{equ:output2}), there can be more than one branches activated in this multi-branch scheme. 
where $\mathbf{K}$ denotes the set of index for $K$ candidate operations. Notably, an identity operation $op_{id}$ is additionally indexed in $\mathbf{K}$ for normal layers with $g_{id}$ always equals 1. 
%\lc{a little justify eg. "notably, the identity operation is excluded form the operation set for reduction layer due to the fact that we impose skip connection during reduction"}

\subsubsection{Weight Update}
%Our preliminary goal is to find a series of architectures that maximize the validation accuracy yet subject to a resource constraint.\yz{refine needed, central: shared weights for fitness evaluation}
%fitness evaluation

%\textbf{Evaluating architectures using weight sharing}. We adopt the weight sharing strategy~\cite{bender2018understanding,liu2018darts,cai2018proxylessnas,guo2019single} to retrieve the validation accuracy for sampled architectures. 

With weight sharing~\cite{bender2018understanding,liu2018darts,cai2018proxylessnas,guo2019single}, every subgraph, \ie architecture denoted by $a$, inherits its weights denoted by  $W_{\mathcal{A}_s}(a)$ from the weights of the supernet denoted by $W_{\mathcal{A}_s}$, where $a  \subseteq \mathcal{A}_s$.
The optimization of shared weights $W_{\mathcal{A}_s}$ can be expressed as follows:
% Instead of training these shared weights by sampling architectures based on a distribution parameterized by jointly optimized importance index $\Theta$ like DARTS \cite{liu2018darts} and ProxylessNAS \cite{cai2018proxylessnas} did, 
% we sample every possible architecture equally based on two consideration: 1. avoid coupling issue during early stage training, which is detrimental to the robustness. 2. we view the shared weight as a good accuracy predictor, only evenly trained supernet can be utilized for such puopose. 
% By sampling architectures from a uniform architecture distribution $\mathbb{U}_{a}$ where every branch path is  sampled with $g \sim Bernoulli(0.5)$, the optimization of shared weights $W_A$ can be expressed as follows:
\begin{equation}
W_{\mathcal{A}_s} =\mathop{\arg\min}_{W_{\mathcal{A}_s}} \mathop{\mathbb{E}}_{a\sim\mathbb{U}_{a}}[\mathcal{L}_{CE}(\mathcal{N}(a,W_{\mathcal{A}_s}(a)))],
\label{eq:WAs}
\end{equation}
where $\mathop{\mathbb{E}}[\cdot]$ denotes expectation, $\mathcal{L}_{CE}(\cdot)$ denotes the cross-entropy loss, and $\mathcal{N}(a,W_{\mathcal{A}_s}(a))$ denotes the network with architecture  $a$ and  parameters $W_{\mathcal{A}_s}(a)$.
%When training the shared weights $W_{\mathcal{A}_s}$ in supernet, typical approaches like DARTS \cite{liu2018darts} and ProxylessNAS \cite{cai2018proxylessnas} sample architectures based on a distribution parameterized by jointly optimized architecture parameter $\Theta$ , which could be regarded as a fitness indicator. However, this strategy has the problem of accumulated advantage, which means the winner of the early-stage takes all, which is harmful to the robustness~\cite{chu2019fairnas}.
%On the other hand,  m
The minimization over expectation $\mathop{\mathbb{E}}[\cdot]$ in Eq.~(\ref{eq:WAs}) is implemented by sampling architecture $a$ from the space $\mathcal{A}_s$  and then updating the corresponding weights $W_{\mathcal{A}_s}(a)$ using stochastic gradient descent.
Multiple works \cite{chu2019fairnas,yu2019evaluating} points out that the supernet needs to be trained evenly so that the shared weights can be good for predicting accuracy.
%Based on the above two considerations,
Therefore, we sample every possible architecture equally, \ie architectures will be sampled from a uniform distribution $\mathbb{U}_{a}$ where every branch is sampled with $g_{n} \sim Bernoulli(0.5)$. 

%
%
%  outline of method revisions:
%  
%   concepts:
%   -fitness evaluation
%   -- individual fitness: from importance indicator
%   -- overall fitness: from validation accuracy
%
% 
%
%
%   replacements:
%   prune --> drop/simplify in method section
%   importance indicator  ---> fitness indicator
%   supernet training  --> fitness evaluation

\subsection{Search Space Evolution}
\label{sec:ias}
Each iteration of search space evolution involves three steps as shown in Figure~\ref{fig:search}, here we provide detailed illustrations for each step.
\subsubsection{Search Space Simplification}
\label{sec:sss}
%When we extend the conventional single-branch structure to multi-branch,  
The multi-branch scheme we adopted enlarge the search space by more than 90 orders of magnitudes (see appendix for detailed numbers). To effectively search for the optimal architecture, we propose to progressively simplify the search space subset with an operation-wise fitness while training the supernet. Specifically, we adopt learnable fitness indicators $\Theta=\{\theta_{n}^{l}\}$ to predict the fitness of an operation and guide the simplification of the search space subset, they are assigned to every operation candidate individually as shown in Figure~\ref{fig:branch}.

%our approach  assigns a fitness indicator $\theta_{\mathcal{I}_l(n_l)}$ for every operation candidate individually.

%instead of assigning a parameter for every possible structure within a learnable layer (which is the combination of the candidate operation choices and will introduce orders of magnitudes more parameters when compared with a single-branch scenario), 

\textbf{Simplifying Search Space with Lock and Rehearse.}
% \label{sec:l&R}
The $n$-th path with its corresponding indicator $\theta_n$ below a certain threshold will be dropped from supernet, except that: 1) the path is the last remained path in a reduction cell, or 2) the operation of this path is inherited from the previous iteration. The second case of exceptions keeps the inherited operations even if their predicted fitness is below the threshold.  It also allows us to re-evaluate previous Pareto optimal architectures on the current supernet, thus protecting the inherited knowledge from being underestimated. This is inspired by the knowledge rehearsal for lifelong learning \cite{silver2002task}. We call this strategy Lock and Rehearse (L\&R).
 
\textbf{Probabilistic Modeling of Fitness Indicators.}
%As shown in~\cite{zhang2020does}, reducing the search space can enable easier training and increase the validation accuracy, which would also lead to a better final result. We follow this insight and progressively simplify the search space 
Fitness Indicators are updated once after every two supernet training iterations. For each update, the $n$-th path is individually sampled with $g_{n} \sim Bernoulli(p_{n})$, where $p_{n} = \frac{1}{1 + e^{-\theta_{n}}}$.
%The values of $\Theta$ will not affect the training of the CNN parameters but only affect the architecture sampling distribution when updating itself. 
%The fitness indicator is learned using simulated gradients (details in Section~\ref{Sec:Objective} and Section~\ref{Sec:Gradient}) and then used for dropping operations.
%\subsubsection{The Objective Function of Fitness Indicators} 
%\label{Sec:Objective}
%  \wl{I recommend change $_i$ into $_{l, n_l}$ to be consistent with the previous definition. Then $g_i$ actually corresponds to $s(l, n_l)$.} 
The fitness indicators $\Theta$ are updated on the validation set as follows:
\begin{equation}
\label{equ:theta}
\Theta^{*} =\mathop{\arg\min}_{\Theta} \mathop{\mathbb{E}}_{a\sim\mathbb{P}_{a}(\Theta)}[\mathcal{L}_{val}(\mathcal{N}(a,W_{\mathcal{A}_s}(a)))],
\end{equation}
where $\mathbb{P}_{a}(\Theta)$ denotes the architecture probability distribution parameterized by fitness indicators $\Theta$. 

% \textbf{Simulated Gradients}
% \label{Sec:Gradient}
% Given a learnable multi-branch cell $C \subset \mathcal{A}$ with $n$ candidate operations, 
For a layer with $K$ candidate operations,
we calculate the layer probability $\hat{p}_{\mathbf{g}}$ for operation configuration $\mathbf{g}$ as the joint probability of $K$-dimensional Bernoulli random variables based on operation probability $p_{n}$:
\begin{equation}
\hat{p}_{\mathbf{g}} = \prod_{n\in\mathbf{K}} (g_{n}p_{n}+  (1-g_{n})(1-p_{n})),\\
\end{equation}
% \begin{equation}
% \begin{array}{rcl}
% \hat{p}_{l}(\mathbf{s}_{l,l_c}) = \prod_{n_l=1}^{N_l} \left( s_{\mathcal{I}_l(n_l)}p_{\mathcal{I}_l(n_l)}+  (1-s_{\mathcal{I}_l(n_l)})(1-p_{\mathcal{I}_l(n_l)})\right).\\
% \end{array}
% \end{equation}
%where $l_c=\sum_{n_l=1}^{N_l}2^{n_l-1}{s_{\mathcal{I}_l(n_l)}}$.
%\yz{where $\mathbf{K}$ denotes the set of index for $K$ candidate operations,}
When initialized, $\theta=0$ for all $\theta \in \Theta$, resulting in $p_{n}=0.5$, so that all possible combinations are selected with equal probability at the beginning. When the fitness indicators $\Theta$ are optimized with resource constraints as regularization, the formulation in Eq.~(\ref{equ:theta}) can be implemented as follows:
\begin{equation}
\begin{aligned}
\label{equ:f}
\Theta^{*} =\mathop{\arg\min}_{\Theta} \mathop{\mathbb{E}}_{a\sim\mathbb{P}_{a}(\Theta)}[&\mathcal{L}_{CE}(\mathcal{N}(a,W_{\mathcal{A}_s}(a))) + \\&\alpha (\mathcal{R}(a, \tau))^{\beta}],
\end{aligned}
\end{equation}
% \begin{equation}
% \label{equ:f}
% %\begin{split}
% \Theta^{*} =\mathop{\arg\min}_{\Theta} \mathop{\mathbb{E}}_{a\sim\mathbb{P}_{a}(\Theta)}\left[\mathcal{L}_{CE}(\mathcal{N}(a,W_{\mathcal{A}_s}(a))) + \alpha (\mathcal{R}(a, \tau))^{\beta}\right]. 
% \end{equation}
where $\mathop{\mathbb{E}}[\cdot]$ denotes expectation, $\mathcal{R}$ measures the difference between the resource demand of architecture $a$ and the target demand $\tau$. %$\hat{c}$ denotes the set of branch combinations in architecture $a$;
$\alpha$ and $\beta$ are application-specific constants. For the latency constraint, we follow \cite{yang2018netadapt} to build up a latency lookup table, which records the latency cost of every operation included in the search space.

\textbf{Simulated Gradients.}
Since $\Theta$ is not directly involved in the computation of $\mathcal{L}_{CE}$, we are unable to update the first term in Eq.~(\ref{equ:f}) directly through back-propagation, thus a simulated gradient is necessary. Inspired by BinaryConnect~\cite{courbariaux2015binaryconnect}, we first forward the output of a sampled branch combination  $\mathbf{o}_{\mathbf{g}_{a}}(\mathbf{X})$ to the next layer, where $\mathbf{g}_{a}$ is a randomly selected configuration. Then we utilize the gradient w.r.t. the output as the simulated gradient.
Hence, we have the simulated gradient written as follows:
\begin{equation}
\label{equ:diff}
\frac{d\mathcal{L}_{CE}}{d\theta_{n}}=\sum_{i=1}^{2^{K}} \frac{d\mathcal{L}_{CE}}{d\hat{p}_{\mathbf{g}_i}} \frac{d\hat{p}_{\mathbf{g}_i}}{d\theta_{n}}\approx \sum_{i=1}^{2^{K}} \frac{d\mathcal{L}_{CE}}{d\mathbf{o}_{\mathbf{g}_{a}}(\mathbf{X})} \mathbf{o}_{\mathbf{g}_{i}}(\mathbf{X}) \frac{d\hat{p}_{\mathbf{g}_i}}{d\theta_{n}},
%= \sum_{j=1}^{l} \frac{d\mathcal{L}_{CE}}{dG_j} \frac{d\hat{p}_{j}}{dp_{i}}\frac{dp_{i}}{d\Theta{i}}
\end{equation}
% \begin{equation}
% \begin{aligned}
% \label{equ:diff}
% \frac{d\mathcal{L}_{CE}}{d\theta_{n}}&=\sum_{i=1}^{2^{N}} \frac{d\mathcal{L}_{CE}}{d\hat{p}_{\mathbf{g}_i}} \frac{d\hat{p}_{\mathbf{g}_i}}{d\theta_{n}} \\ 
% &\approx \sum_{i=1}^{2^{N}} \frac{d\mathcal{L}_{CE}}{d\mathbf{o}_{\mathbf{g}_{a}}(\mathbf{X})} \mathbf{o}_{\mathbf{g}_{i}}(\mathbf{X}) \frac{d\hat{p}_{\mathbf{g}_i}}{d\theta_{n}},
% %= \sum_{j=1}^{l} \frac{d\mathcal{L}_{CE}}{dG_j} \frac{d\hat{p}_{j}}{dp_{i}}\frac{dp_{i}}{d\Theta{i}}
% \end{aligned}
% \end{equation}
where $\mathbf{o}_{\mathbf{g}_{i}}(\mathbf{X})$ is defined in Eq.~(\ref{equ:output2}).
%Following  \cite{cai2018proxylessnas}, 
The summation over all $\mathbf{g}_{i}$ in Eq.~(\ref{equ:diff}) is still complex. We use the method in~\cite{cai2018proxylessnas} to simplify the computation by selecting only two configurations to reduce the GPU memory and computation required for each iteration.
%, we provide further details in the appendix
%We further simplify the computation and select only two configurations.
Specifically, we randomly sample anther configuration $\mathbf{g}_{b}$, and then rescale the layer probabilities $\hat{p}_{\mathbf{g}}$ of configuration $\mathbf{g}_{a}$ and $\mathbf{g}_{b}$ to $\tilde{p}_{\mathbf{g}}$ so that $\tilde{p}_{\mathbf{g}_{a}} + \tilde{p}_{\mathbf{g}_{b}} = 1$.
Finally, we have the simulated gradient approximated as:
\begin{equation}
\begin{aligned}
\label{equ:valloss}
\frac{d\mathcal{L}_{CE}}{d\theta_{n}} \approx & \frac{d\mathcal{L}_{CE}}{d\mathbf{o}_{\mathbf{g}_{a}}(\mathbf{X})} \mathbf{o}_{\mathbf{g}_{a}}(\mathbf{X}) \frac{d\tilde{p}_{\mathbf{g}_{a}}}{d\theta_{n}} + \\ &\frac{d\mathcal{L}_{CE}}{d\mathbf{o}_{\mathbf{g}_{a}}(\mathbf{X})} \mathbf{o}_{\mathbf{g}_{b}}(\mathbf{X}) \frac{d\tilde{p}_{\mathbf{g}_{b}}}{d\theta_{n}}.
\end{aligned}
\end{equation}
% \begin{equation}
% \begin{aligned}
% \label{equ:valloss}
% \frac{d\mathcal{L}_{CE}}{d\theta_{n}} \approx & \frac{d\mathcal{L}_{CE}}{d\mathbf{o}_{\mathbf{g}_{a}}(\mathbf{X})} \mathbf{o}_{\mathbf{g}_{a}}(\mathbf{X}) \frac{d\tilde{p}_{\mathbf{g}_{a}}}{d\theta_{n}} +  \\ &\frac{d\mathcal{L}_{CE}}{d\mathbf{o}_{\mathbf{g}_{a}}(\mathbf{X})} \mathbf{o}_{\mathbf{g}_{b}}(\mathbf{X}) \frac{d\tilde{p}_{\mathbf{g}_{b}}}{d\theta_{n}}.
% \end{aligned}
% \end{equation}
Likewise, the $\mathcal{R}(a, \tau)$ of second term in Eq.~(\ref{equ:f}) is approximated as:
\begin{equation}
% \begin{aligned}
\label{equ:ff}
\mathcal{R}(a, \tau) \approx  \sum_{l=1}^{L}(\tilde{p}_{\mathbf{g}_{a}^{l}}\mathcal{C}(\mathbf{o}_{\mathbf{g}_{a}^{l}}) + \tilde{p}_{\mathbf{g}_{b}^{l}}\mathcal{C}(\mathbf{o}_{\mathbf{g}_{b}^{l}})) - \tau,
% \end{aligned}
\end{equation}
where $\mathcal{C}$ maps the combination of operations $\mathbf{o}_{\mathbf{g}}$ to its corresponding resource cost.
%--------------------------------------------
\subsubsection{Pareto Front Retrieval and Architecture Aggregation}
\label{sec:pofront}
%Previous approaches directly take the architecture with the highest importance indicators as the final optimal result.
%We argue that such an architecture is only a statistically learned importance indicator, yet 
%As the NAS problem by its nature is a discrete combinatorial optimization problem, where each learnable unit does not affect the overall performance independently.
%\wl{independent assumption (what is this assumption?)} does not hold.
%Simply selecting a single architecture according to the individual indicators may miss those high-performance architectures with better operation combinations. \wl{if it is motivation, put it in the introduction. If it is analysis, put it after introducing the method.}
%\wl{On the other hand, one-shot based methods~\cite{bender2018understanding,lu2018nsga} propose to take the supernet as an accuracy predictor.  Their results are retrieved by Evolutionay algorithms, and a single model or Pareto-frontier is obtained finally.(what is the point of mentioning this?)}

% if resource constraints are considered.
\textbf{Pareto Front Retrieval.}
To retrieve the Pareto front needed for architecture aggregation, we evaluate the validation accuracies of sampled architectures by using the weights of the well-trained supernet and the validation dataset.
%As the search space of the supernet has been well simplified to a relatively small size upon Pareto front retrieving, 
We randomly sample $D$ distinct models based on the probability distribution implied by fitness indicators $\Theta$. The sampled models not fitting the resource constraint are discarded. The Pareto-optimal architectures from the last searching iteration, if exist, will also be evaluated as part of the L\&R strategy.
In practice, extra $D_e$ samples are needed to overcome the edging effect (see appendix for details).
%\yz{$D=2000, D_e=100$ been moved to appendix}
After all samples are evaluated, we can get $P$ Pareto-optimal architectures $\{a_1, \ldots, a_p, \ldots, a_P\}$. The Pareto-optimal architectures derived from the final round of optimization will be referred to as final results.

Due to the volatility of BN statistics for supernet and the fact that different architectures should adopt different BN statistics, we need to perform recalculation for BN layers.
Specifically, before evaluating an architecture on the shared weight supernet, we recalculate the statistics of BN layers by forwarding 20k random training images, which takes about one second.

% \subsubsection{Search Space Aggregation and Inheritance}
% \label{sec:agg&inher}
\textbf{Aggregation.}
After Pareto front retrieval, we take the union of operations from all $P$ Pareto-optimal architectures to get the optimized search space $\hat{\mathcal{A}_s}$.
Mathematically, we denote $\mathbf{e}_{l}^{p} = \{op_{n}^{l}|g_{n}^{l}=1, n\in\mathbf{K}_{l} \}$ as the selected operations of $l$-th layer for the $p$-th  Pareto-optimal architecture $a_p$, and denote $\hat{\mathbf{E}}_{l}^{s}$ as the optimized search space subset of the layer $l$ in $\hat{\mathcal{A}_s}$.
%For the union of $P$ architectures, the operations of layer $l$ across all $P$ architectures will be collected as the optimized search space for layer $l$.
We have $\hat{\mathbf{E}}_{l}^{s} = \bigcup_{p=1}^{P}\mathbf{e}_{l}^{p}.$ 
%represented as follows:
% \begin{equation}
% % \hat{\mathcal{A}_s} = (\bigcup\limits_{i=1}^{P} a_{i}).
% \hat{\mathcal{A}_s^{l}} = \bigcup\limits_{p=1}^{P} \{op_{\mathcal{I}_l(n_p)}|s_{\mathcal{I}_l(n_p)}=1\}_{n_p=1}^{K}.
% \label{eq:AShat}
% \end{equation}

\subsubsection{Inheritance by Search Space Replenishment.}
\label{sec:Inher}
To obtain the new search space subset $\mathcal{A}_{s}'$ for the next round of evolution, we randomly sample a certain amount of candidate operations (excluding previoulsy traversed operations) in the overall search space pool $\mathcal{A}$ as the supplemental operations $\mathcal{A}_{r}$.  Besides, the optimized search space subset $\hat{\mathcal{A}_s}$ obtained by aggregation %in Eq.~(\ref{eq:AShat}) 
will be inherited.
%The $\mathcal{A}_{r}^l$ will replenish the search space subset 
Specifically, we have $\mathcal{A}_{s}' = \hat{\mathcal{A}}_s \cup \mathcal{A}_{r}$, where the new search space subset $\mathcal{A}_{s}'$ is the union of the  supplemental operations $\mathcal{A}_{r}$ and the optimized search space subset $\hat{\mathcal{A}}_s$. After replenishment, the size of the resulting search space subset $\mathcal{A}_s'$ will be the same as the size of the original $\mathcal{A}_s$ with $K$ candidate operations per layer. Finally we have $\mathcal{A}_s = \mathcal{A}_s'$. 

%-------------------------------------------------------------------------
\section{Experimental Results}
%To demonstrate the effectiveness of proposed method, we report experimental results for NSE under ImageNet mobile settings with two type of constraints in Section \ref{sec:FLOPs} and Section \ref{sec:lat} respectively. We then compare NSE with existing techniques to investigate the necessity of proposed approach under large search space in Section \ref{sec:Necessity}. 
\begin{table}[t]
    \footnotesize
    \begin{center}

        \begin{tabular}{p{3.5cm}|>{\centering}p{1cm}|>{\centering}p{1cm}|>{\centering}p{1cm}}
            % \hline
            Network            & Params &FLOPs    & Top-1     \tabularnewline
            % \hline\hline
            \hlineB{2.5}
            MobileNetV2 1.4~\cite{sandler2018mobilenetv2}        & 6.9M     &585M    &74.7     \tabularnewline
            ShuffleNetV2 2$\times$~\cite{ma2018shufflenet}      & -     &591M     &74.9         \tabularnewline

            NASNet-A~\cite{zoph2018learning}         &5.3M    &564M    &74.0         \tabularnewline
            DARTS~\cite{liu2018darts}         &4.7M    &574M    &73.3             \tabularnewline
            Proxyless-mobile~\cite{cai2018proxylessnas}   & 4.1M &320M     &74.6             \tabularnewline
            FBNet-C~\cite{wu2018fbnet}            &5.5M    &375M        &74.9        \tabularnewline
            %BetaNAS &  3.6M&315M&75.1&92.3\\
\hline
            NSENet-27                  &4.6M&325M    &75.3        \tabularnewline
            \textbf{NSENet}                  &\textbf{4.6M}&\textbf{330M}    &\textbf{75.5}        \tabularnewline
            \hline
            \hline
            ShuffleNetV2 2$\times$~\cite{ma2018shufflenet}$\dagger$  & -     &597M     &75.4     \tabularnewline
            % MnasNet-A1~\cite{tan2018mnasnet}$\dagger$                    & 3.9M    &312M    &75.2          \\
            MnasNet-A2~\cite{tan2018mnasnet}$\dagger$                    & 4.8M    &340M    &75.6          \tabularnewline

            %BetaNAS 1.4$\times$$\dagger    \ddagger$&7.2M&631M&79.0&94.2\\
            MixNet-S~\cite{tan2019mixconv}$\dagger$&4.1M&256M&75.8\tabularnewline
            MixNet-M~\cite{tan2019mixconv}$\dagger$&5.0M&360M&77.0\tabularnewline
            % MobileNetV3-large~\cite{howard2019searching}$\dagger$ &5.4M&219M&75.2\\
            MobileNetV3-large/1.25~\cite{howard2019searching}$\dagger$&7.5M& 356M&76.6\tabularnewline
            GreedyNAS-B~\cite{you2020greedynas}$\dagger$&5.2M&324M&76.8\tabularnewline
            % GreedyNAS-A~\cite{you2020greedynas}$\dagger$&6.5M&366M&77.1\\
            %MixNet-L&7.3M&565M&78.9&94.2\\

            %SE-DARTS+$\dagger    \ddagger$  &6.1M&594M&77.5&93.6 \\
            EfficientNet-B0~\cite{tan2019efficientnet}$\dagger    \ddagger$ &5.3M&390M&76.3 \tabularnewline
            % \toadd{AtomNAS-B+} \cite{mei2019atomnas} $\dagger \ast$ &5.5M&329M&77.2 \\
            % \toadd{FBNetV2-L1} \cite{wan2020fbnetv2}$\dagger \ast$ &-&325M&77.2 \\
            %EfficientNet-B1$\dagger    \ddagger$ &7.8M&700M&78.8&94.4 \\

            \hline
            \textbf{NSENet}$\dagger$ &\textbf{7.6M}&\textbf{333M}    &\textbf{77.3} \tabularnewline
            % \toadd{\textbf{BufferNet}} $\dagger\ast$ &xM&xxxM    &xx.x \\
            
            %\remove{\textbf{BufferNet} $\dagger \ddagger$ \alert{To Be Removed}} &\remove{7.6M} &\remove{333M} &\remove{77.1} \\
            % 330 54
            % \hline
            %Ours(latency) + SE,HSwish,Mixup              & 19.90M&9.1 + 1 ms?    &78.83            \\ % 340
            %\hline

    \end{tabular}
\end{center}
    \caption{ImageNet results compared with state-of-the-art methods in the \textit{mobile} setting. NSENet-27 denotes the network we discovered in the 27 OPs space. NSENet denotes the network found by exploring second space based on the search space subset we achieved from the 27 OPs space.  $\dagger$ denotes model using extra modules such as swish activation~\cite{ramachandran2017searching} and SE module~\cite{hu2018squeeze}. $\ddagger$ denotes model trained with AutoAugment \cite{zoph2019learning}.
    %, $\ddagger$ 
    %\toadd{$\ast$ denotes model searched with fine-grained search space (including backbone channel, op channel and resolution).}
    }
    % \vspace{-3.5mm}
\label{tab:FLOPs}
\end{table}

\subsection{Search Space and Configurations}
\label{sec:FLOPsSpace}

\textbf{Search Space.}
%The overall architecture of our search space is similar to Proxyless \cite{cai2018proxylessnas} and is consists of multiple building blocks (see appendix).
For FLOPs constrained experiments, our primary search space consists of 27 distinct operations (27 OPs space). By inheriting the final search space subset derived from the 27 OPs space, we continue to search for three extra rounds over a complete new search space (second space) with OPs not covered in 27 OPs space to get the final result.
For Latency constrained experiments, we use the 19 operations search space (19 OPs space)~\cite{li2020improving}. Notably, the number of OPs is quite large compared with \cite{liu2018darts,cai2018proxylessnas}.
The detailed search space and structure of searched architectures can be found in the appendix. 

The order of the candidate operations is randomly shuffled layer-wisely to minimize the potential influence from its ordering, yet all our experiments share the same set of shuffled sequences for a fair comparison. As shown in Figure~\ref{fig:search}, there could be an imbalanced number of operations being preserved in each layer, which means that some layers could be running out of candidates ahead of time. We bypass this issue by taking the moment when such a shortage happened as the end of an NSE search process.

%Specifically, we load the optimized sub search space derived from the previous space and then continue to search over a space consists of the grouped DW convolution variant that appeared in MixNet \cite{tan2019mixconv} for 3 extra rounds.

%\toadd{ToADD: } For fine-grained model adjustments, we first fix previously searched building blocks, then add two extra options for input resolution, backbone channels and operation channels, with their default values moved up or down by $14\%, 20\%,  20\%$ respectively. We share the needed weights for different options like what FBNetV2 \cite{wan2020fbnetv2} has done and train the supernet with uniform sampling but without pruning (see appendix for details).

\textbf{Configurations.}
The layer-wise size of the search space subset $K$ is set to 5 for FLOPs constraint and 6 for Latency constraint. To get the final model, we randomly sample 5 of the final Pareto optimal points and rescale them to approximate 330M FLOPs, the top-performing model after retraining is selected as the output model (see appendix for detailed setups). All experiments are performed on the ImageNet \cite{ILSVRC15} dataset, where the validation set is constructed by 50K random images sampled from the training set.

%For every supernet training, we use Nesterov SGD with  0.9 momentum, weight decay $4e^{-5}$, batch size 1024 with 100 epochs. The initial learning rate is 0.4 and gradually reaches 0 through cosine learning rate decay. We use Adam optimizer with an initial learning rate of 0.1 to update importance indicator, and we perform such update every two supernet updates.
%The pruning threshold is set to -2.
%For model retraining, we increase the number of epochs to 350, with batch size 2048, learning rate 0.8 and exponential moving average with decay 0.9999. For a fair comparison, swish activation, SE modules together with identical training configs from EfficientNet \cite{tan2019efficientnet} are optionally used subject to the specific settings as shown in Table \ref{tab:FLOPs}.

\begin{figure*}[t]
    \centering
    \subfigure[]{
    \centering
    \includegraphics[width=0.32\linewidth]{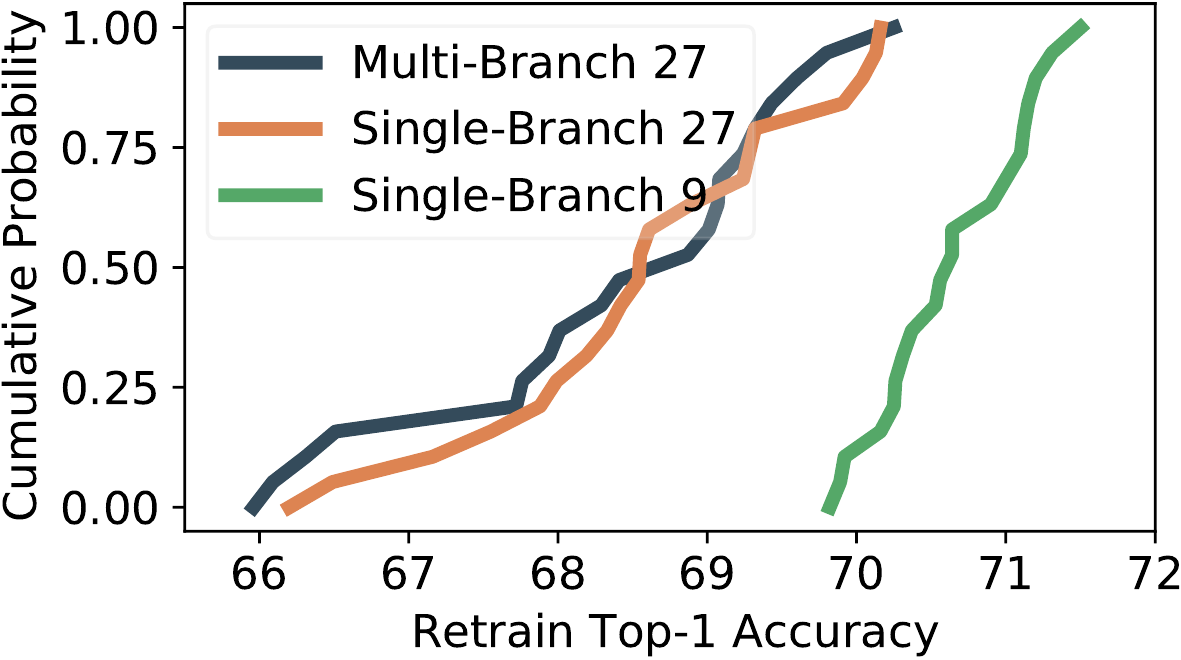}}
    % \scalebox{0.4}{\input{random.png}}}
    \label{CRFsearchspace}
    \subfigure[]{
    \centering
    \includegraphics[width=0.32\linewidth]{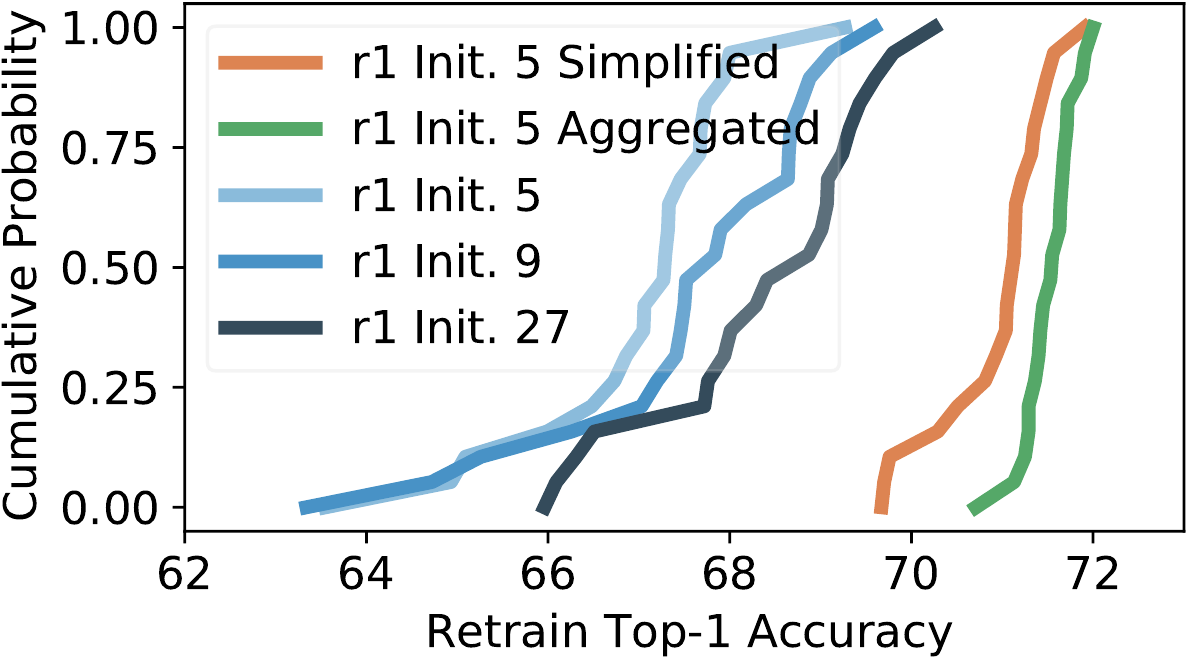}}
    % \scalebox{0.4}{\input{reduce.png}}}
    \subfigure[]{
    \centering
    \includegraphics[width=0.32\linewidth]{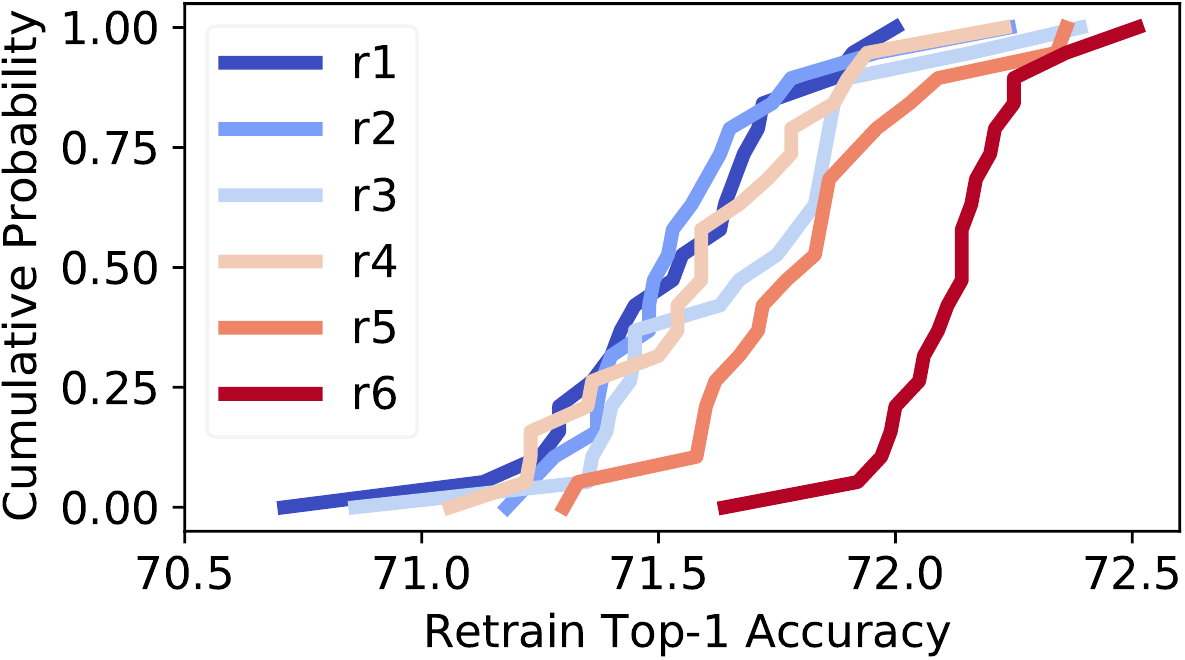}}
    % \scalebox{0.4}{\input{convergence.png}}}
        \label{prunespace}
    \caption{Search space comparisons conditioned on FLOPs. For each experiment, we randomly sample 20 architectures that have FLOPs within the interval of [323M, 327M]. Each model is then trained from scratch for 50 epochs. (b): r1 init $n$ denotes the randomly initialized search space with layer-wise space size $K=n$, ``Simplified" and ``Aggregated" respectively denote the remained search space after simplification and the search space achieved by Pareto front aggregation. (c): r$n$ denotes the search space derived from the aggregation of Pareto optimal in the $n$-th round. The experiment performed in (b) and (c) are both based on the Multi-Branch 27 OPs space.}
    \vspace{-2.mm}
        \label{fig:crf}
\end{figure*}

\subsection{FLOPs Constraint Results}
\label{sec:FLOPs}

The target $\tau$ for FLOPs constraint is 300M FLOPs. As shown in Table \ref{tab:FLOPs}, our preliminary result (NSENet-27 in Table \ref{tab:FLOPs}) derived from the 27 OPs space achieves 75.3\% Top-1 accuracy with 325M FLOPs, which has already surpassed many of the hand-crafted or automatically designed architectures. When our search space subset continues to accommodate the second space, the result (NSENet in Table \ref{tab:FLOPs}) further pushes the top performance of the derived model to 75.5\% Top-1 accuracy. When auxiliary techniques are considered, our model (\textbf{NSENet}$\dagger$ in Table \ref{tab:FLOPs}) consistently surpass previous models by a considerable margin.

\subsection{Latency Constraint Results}
\label{sec:lat} %!!!

\begin{table}[t]
    \footnotesize
    \begin{center}
        \begin{tabular}{p{3.5cm}|>{\centering}p{1cm}|>{\centering}p{1cm}|>{\centering}p{1cm}}
            % \hline
            Network           &Params  &Latency$^\ast$    & Top-1    \tabularnewline
            \hlineB{2.5}
            MobileNetV2 1.4~\cite{sandler2018mobilenetv2} &6.9M        &8.9 ms    &74.7             \tabularnewline
            ShuffleNetV2 2$\times$~\cite{ma2018shufflenet} &-          &6.8 ms     &74.9             \tabularnewline
            \hline
            NASNet-A~\cite{zoph2018learning}   &5.3M      &23 ms    &74.0             \tabularnewline
            % AmoebaNet-A~\cite{real2018regularized}                 &33 ms    &74.5             \tabularnewline
            PNASNet~\cite{liu2018progressive}    &5.1M            &25 ms     & 74.2             \tabularnewline
            Proxyless-GPU~\cite{cai2018proxylessnas}  &7.1M  &7.9 ms     &75.1            \tabularnewline
            PC-NAS-L~\cite{li2020improving}   &15.3M &10.3 ms     &77.5            \tabularnewline
            %EfficientNet-B0$\dagger \ddagger$ &about 13ms&76.3 \\
            % MnasNet~\cite{tan2018mnasnet}$\dagger$                        &11 ms    &74.8             \\
            MixNet-S~\cite{tan2019mixconv} $\dagger$ &4.1M &27 ms&75.8\tabularnewline
            MobileNetV3-large/1.25~\cite{howard2019searching} $\dagger$&7.5M &10.5 ms &76.6\tabularnewline
            \hline
            \textbf{NSENet-GPU}               &\textbf{15.7M}  &\textbf{8.9 ms}    &\textbf{77.9}            \tabularnewline
            % 330 54
            % \hline
            %Ours(latency) + SE,HSwish,Mixup              & 19.90M&9.1 + 1 ms?    &78.83            \\ % 340
            %\hline

        \end{tabular}
    \end{center}
    \caption{ImageNet results compared with state-of-the-art methods in the \textit{latency constrained} setting. NSENet-GPU is the network we found with the 19 OPs space \cite{li2020improving}. $\dagger$ denotes model using extra modules such as swish activation \cite{ramachandran2017searching} and SE module \cite{hu2018squeeze}. $\ast$ latency for all networks are evaluated under the same setting (batch size 16 on GTX TITAN Xp GPU and TensorRT3 framework).
    }
    % \vspace{-3.5mm}
\label{tab:LATs}
\end{table}

The platform we optimize for is GTX TITAN Xp GPU and TensorRT3 framework. All the latency was evaluated with the batch size set to 16 to fully utilize the GPU resource. The resource constraint is targeted at 8 ms.
The result is shown in Table \ref{tab:LATs}, our NSENet-GPU obtains $77.9\%$ top1 accuracy with 8.9 ms latency cost.  Notably, while we share the identical search space with PC-NAS-L (see appendix), our searched model performs $0.4\%$ better in terms of Top-1 accuracy with 1.4 ms less latency cost.

\subsection{Necessity of Search Space Evolution}
\label{sec:Necessity}
 %This section gives detailed results.
\begin{table}[t]
\footnotesize
% \vspace{-11mm}
\begin{center}
    \begin{tabular}{p{2.7cm}|>{\centering}p{1.8cm}|>{\centering}p{1cm}|>{\centering}p{1cm}}
    % \hline%\noalign{\smallskip}
    Network(s) & Search Cost $\dagger$ & FLOPs & Top-1\tabularnewline
    %\noalign{\smallskip}
    \hlineB{2.5}
    %\noalign{\smallskip}
    Proxyless-mobile~\cite{cai2018proxylessnas} & 200 $\ddagger$ & 320M & 74.6 \tabularnewline
    Proxyless~\cite{cai2018proxylessnas}-rand6 & 2200$^\ast$ & 327M$^\ast$ & 74.2$^\ast$ \tabularnewline
    \hline
    Proxyless~\cite{cai2018proxylessnas}-27  & 4,000 & 336M & 74.5 \tabularnewline
    One-Shot~\cite{bender2018understanding}-27 & 4,000 & 339M & 73.5 \tabularnewline
    \textbf{NSENet-27 $K=5$} & \textbf{4,000} & \textbf{327M}${}^\ast{}^\ast$  &\textbf{75.0}${}^\ast{}^\ast$ \tabularnewline
    % \hline
    \end{tabular}
\end{center}
    \caption{Comparison with different algorithms under 27 OPs space. $\dagger$ search cost refers to GTX 1080Ti GPU hours, $\ddagger$ number cited from \cite{cai2018proxylessnas}, which uses GTX V100 GPU.  $^\ast$~numbers are the average over 3 runs. $\ast\ast$ numbers are the average of final results shown in Figure~\ref{fig:pointobs}(a) 'K=5'.
    The 27 OPs space contains the search space used by Proxyless-mobile and Proxyless-rand6. 
    %The results for Proxyless-mobile are from the paper in ProxylessNAS~\cite{}.
    Proxyless-mobile~\cite{cai2018proxylessnas}, Proxyless-rand6, and Proxyless-27 are searched using the ProxylessNAS~\cite{cai2018proxylessnas} algorithm. % removed to save space
    } % \ask
    \vspace{-2.mm}
\label{table:space_cost}
\end{table}
\textbf{Existing Methods on Large Search Space.}
In the introduction, we conclude that existing methods cannot deal with large search space effectively, which is our motivation. 
%To investigate if we can address the large search space by existing methods, 
We assess 4 reasonably fast methods: DARTS~\cite{liu2018darts}, One-Shot~\cite{bender2018understanding}, SPOS~\cite{guo2019single} and ProxylessNAS~\cite{cai2018proxylessnas}. % as baselines.% (Non-efficient methods \cite{tan2018mnasnet,} failed to converge in competitive time).
%The baselines is applied with longer training epochs to match the gpu hours of NSE. % These results show that an enlarged operation search space increases the difficulty of architecture search and this difficulty can not be resolved simply by a higher search cost.
%Large space brings challenges. 

Algorithms like DARTS \cite{liu2018darts} retain all paths for optimization, making its search cost for our large 27 OPs space prohibitively high (about 27k GPU hours for 100 epochs). Simple one-shot algorithms like SPOS~\cite{guo2019single} do not retain all paths but fail to converge reasonably under the same setting we used.
%, which hurts the search quality. 
Specifically, the supernet trained using SPOS \cite{guo2019single} on our 27 OPs space does not converge (~1\% Top-1 accuracy for 100 epochs), even after the attempts to tune the batch size, learning rate, or gradient clips. 

With the one-shot strategy described in \cite{bender2018understanding}, the supernet trained on our 27 OPs space also converges poorly even after a training schedule that costs 4k GPU hours (aligned by increasing training epochs). The sampled Pareto-optimal architectures with 300M-350M FLOPs have Top-1 validation accuracies less than 30\% on supernet and the final result is 73.5\% Top-1 accuracy as shown in Table~\ref{table:space_cost}.

Algorithms like ProxylessNAS~\cite{cai2018proxylessnas} utilizes learnable architecture parameters to gradually narrow the search space, which helps to get a more reasonable result under a large search space. We run ProxylessNAS on our 27 OPs space and aligned search cost with NSE by increasing the number of training epochs, result denoted by Proxyless-27, is shown in Table~\ref{table:space_cost}. We can observe that ProxylessNAS produces a better result than One-Shot-27, but NSE still 
performs better than ProxylessNAS.

\textbf{ProxylessNAS with Different Search Spaces.}
We choose the previous NAS algorithm with the highest 27 OPs space result compared in Table~\ref{table:space_cost}, ProxylessNAS, to further demonstrate how search space affects its performance. First, we compare its original result Proxyless-mobile with Proxyless-27. Proxyless-mobile is derived from a manually designed search space in ProxylessNAS~\cite{cai2018proxylessnas} with only 6 OPs, a subset of our 27 OPs space. Nevertheless, its accuracy is even slightly higher than Proxyless-27, which means ProxylessNAS is not as effective in 27 OPs space as in 6 OPs space. 
Then we compare it with the result on random 6 OPs subsets, denoted as Proxyless-rand6. The random subsets are constructed by randomly sampling 6 OPs from our 27 OPs space for each layer of the whole search space. The result shows that the random subset space is not as good as the manually designed 6 OPs space or the full 27 OPs search space.
Noticing that the search cost for Proxyless-rand6 is significantly higher than Proxyless-mobile even considering different GPU used, this is caused by less time-efficient OPs within the random subset.
%Which is consistent with our observation in Figure~\ref{fig:crf}(a) and \ref{fig:crf}(b). 

\begin{figure}[t]

    \centering
    \subfigure[]{
    \centering
    \includegraphics[width=0.66\linewidth]{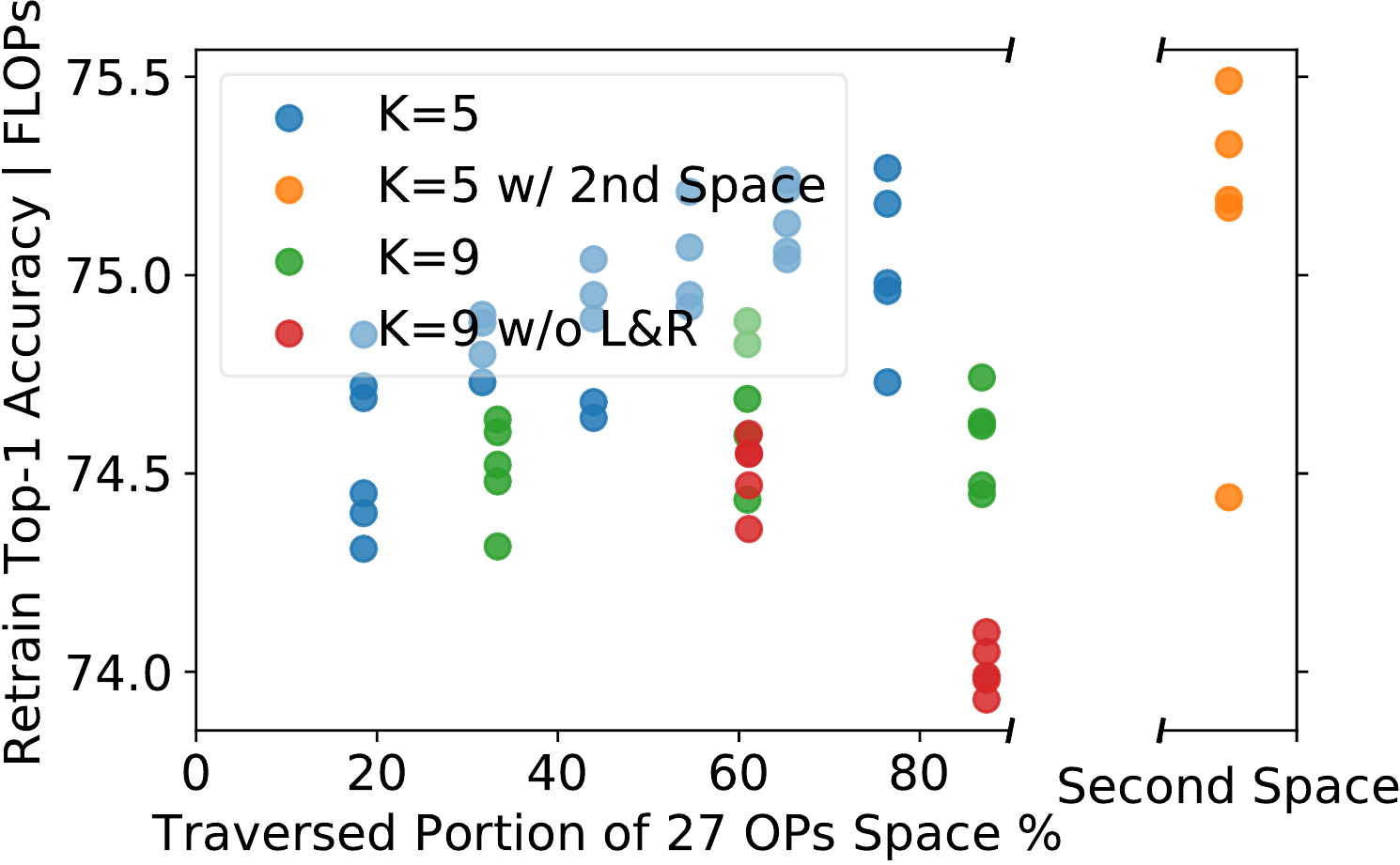}}
    % \scalebox{0.3}{\input{order.pdf_tex}}}
    \subfigure[]{
    \label{lifelong_nas}
    \centering
    \includegraphics[width=0.29\linewidth]{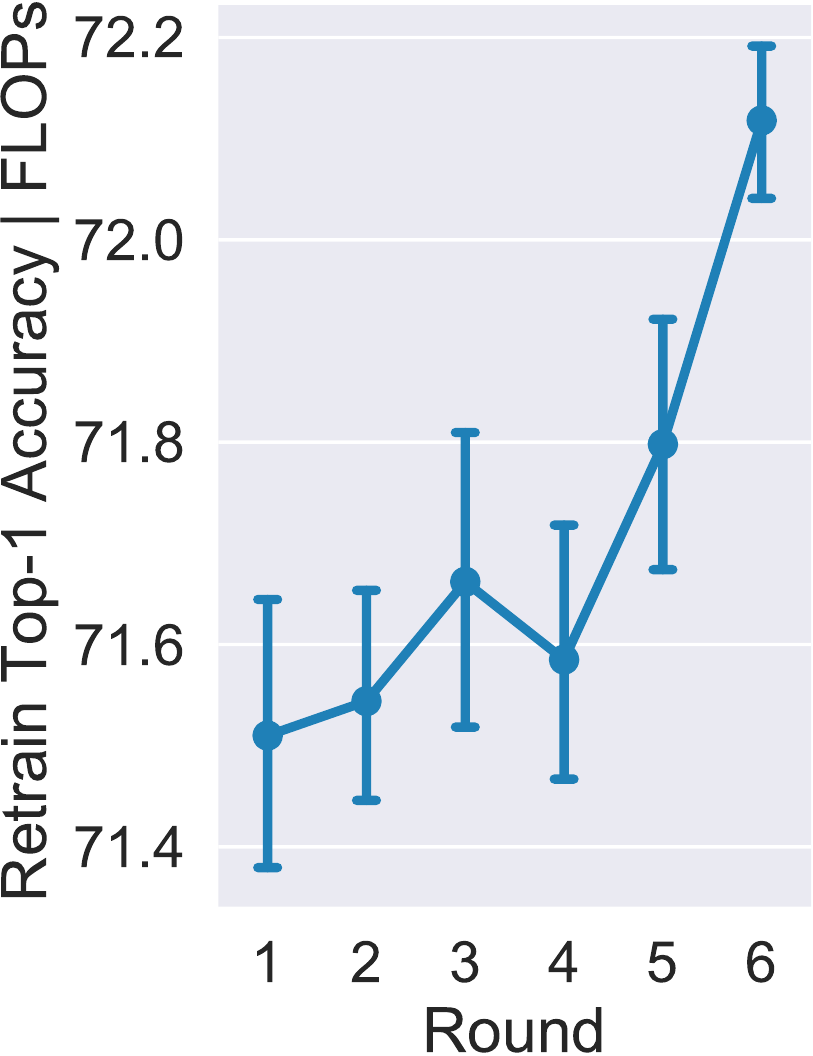}}
    % \scalebox{0.3}{\input{conver.pdf_tex}}}
    \caption{Convergence pattern along the NSE optimization trajectory. (a) Point evaluations of intermediate Pareto optimal points. Models are fully trained from scratch. The entire search process does not cover the full search space because of the inequity issue as mentioned in  Section \ref{sec:FLOPsSpace}.   (b) The progressive improvement of aggregated search space as search proceeds. The plot shares the same set of data with Figure~\ref{fig:crf}(c) and is shown in mean accuracy with 95\% confidence intervals.}
    \vspace{-2.mm}
    \label{fig:pointobs}
\end{figure}

\textbf{Search Space Simplification Without  Search Space Evolution.}
We also investigate whether search space simplification in Section~\ref{sec:sss} alone can handle large search space. %looked into our weight-sharing based NAS algorithm aided with search space simplification and see if it can handle a large search space well. 
We compare the cases of having layer-wise space size $K=5$ and $K=9$, which correspond to `K=5' and `K=9' in Figure~\ref{fig:pointobs}(a) respectively. Note that the initial search space for $K = 5$ is a subset of the initial space used by $K = 9$. 
The first round of search results for $K = 5$ and $K = 9$ in Figure~\ref{fig:pointobs}(a) shows that a larger initial search space (K=9) leads to an inferior result when compared with a smaller one (K=5), although both experiments enable search space simplification. 
%Noticing that both experiments have their initial search space taken from an identically shuffled operation sequence as mentioned in Section \ref{sec:FLOPsSpace}, which means that 
The results show that search space simplification without evolving search space cannot handle large search space well.
%, such subset search space is of lower quality as shown in Figure \ref{fig:crf}(b).
%, nevertheless, a larger space does not perform well enough solely with the help of .

%-------------------------------------------------------------------------
\section{Ablation Studies}

\subsection{Comparison of Search Space Quality}

%A larger search space would cover more candidate operations thus lifting the theoretical upper-bound of the final result. However, it can also make architecture search more difficult. 
To have further insight into whether a search space is competitive for the specific task, we adopt the distribution estimate \cite{radosavovic2019network}, \ie the distribution of retrain accuracy derived from architectures randomly sampled within a search space, to evaluate search space quality.
%of search space quality draw random architecture samples from the search space and use the distribution of their retrain accuracy as a .
%To characterize the quality of a space, %We can further 
Figure~\ref{fig:crf}(a) compares three search spaces to discuss the potential influence of enlarging search space in our experiments. 
The Multi-Branch 27 is the space we used for our key results, Single-Branch 27 stands for the single-branch version of our 27 OPs space, and the Single-Branch 9 is a subset of our 27 OPs space: only DW convolutions with kernel size $\{3, 5, 7\}$ and  expand ratio $\{1, 3, 6\}$, a commonly used group of operations for NAS~\cite{cai2018proxylessnas, wu2018fbnet, you2020greedynas}, is included.
% search space consists of depthwise convolution branches with kernel size $\{3, 5, 7\}$ and  expand ratio $\{1, 3, 6\}$, which are the most  architecture search \cite{cai2018proxylessnas,wu2018fbnet,tan2019mixconv,howard2019searching,guo2019single} and part of our Multi-Branch 27 search space.
It can be seen that there is no outstanding gap between the single-branch and multi-branch space. Besides, while the Multi-Branch 27 search space theoretically has a higher upper bound for potential architectures, the single-branch 9 space, its subset, has a significantly better architecture distribution in terms of retrain accuracy. 
This gap, however, is the very issue we want to address through search space evolution.
%We can also see that there is no significant gap between single-branch and multi-branch structures, this might indicate that on average, multi-branch architectures are as competitive as their single branch counterpart.

\subsection{Continual Convergence}
To demonstrate that our approach can continuously improve the search space subset as well as the searched architectures, 
%we evaluate the retrain accuracy of Pareto-optimal architectures derived from each round of search space evolution.  
we first randomly select 5 Pareto-optimal architectures captured by every round of evolution on the preliminary space, then train them from scratch. As shown in Figure~\ref{fig:pointobs}(a), a continuous trend of improvement in terms of the upper bound and the distribution of results is observed. 
The best model for the entire process can be achieved by inspecting the most recent Pareto-optimal architectures. 
Furthermore, the quality of the aggregated search space has also been progressively improved as shown in Figure~\ref{fig:crf}(c) and \ref{fig:pointobs}(b). After the suspension of an NSE process, we can still restart the process by reusing the final optimized search space subset and replenishing it with a second search space. In this way, our approach (`K=5 w/ 2nd Space' in Figure~\ref{fig:pointobs}(a)) consistently achieves a gain.

\subsection{Impact of Components}
\textbf{Layer-wise Size of Search Space Subset.} As shown in Figure~\ref{fig:crf}(b), comparing layer-wise space size $K = 5$ (`r1 Init 5' in \ref{fig:crf}(b)) and $K = 9$ (`r1 Init 9' in \ref{fig:crf}(b)), smaller $K$ has a poorer accuracy distribution initially. However, a smaller $K$ can still lead to a better convergence as shown by the comparison between `K=5' and `K=9' in Figure~\ref{fig:pointobs}(a). This is coherent with our motivation to bypass the difficulty of large search space by progressive search space evolution, as a smaller search space is easier to optimize.
%due to the fact that expanding the search space would increase the difficulty for NAS.

\textbf{Simplification and Aggregation.} To show the effectiveness of our search space simplification and aggregation, we plot the quality of a subset search space, together with its simplified and aggregated space in Figure~\ref {fig:crf}(b). It shows that both search space simplification (`r1 Init 5 Simplified' in the figure) and search space aggregation (`r1 Init 5 Aggregated' in the figure) improve the search space quality by a considerable margin.

\textbf{Lock and Rehearse.}
Another mandatory procedure in our pipeline is
%to address the {\it catastrophic forgetting} issue. We adopted 
the Lock and Rehearse strategy, which prevents the inherited search space from being underestimated. The absence of such a regularization method could lead to a significant performance decrease as shown by the result `K=9 w/o L\&R' in Figure~\ref{fig:pointobs}(a).

%-------------------------------------------------------------------------
\section{Conclusions}
In this paper, we have introduced a new neural architecture search scheme called NSE. It
targets large space architecture search by progressively accommodates new search space while maintaining the previously obtained knowledge. We further extended the flexibility of obtainable architectures by introducing a learnable multi-branch setting. %We achieved 77.1\% top-1 accuracy on ImageNet with only 333M FLOPs, which set a new state-of-the-art among  auto-generated architectures. By introducing the latency constraint, our result also outperformed previous mobile setting models with a 77.9\% Top-1 accuracy.
Our proposed NSE scheme provides a consistent performance gain with a stream of incoming search spaces, which has minimized the necessity of search space engineering and leads to a step towards fully automatic neural architecture search. \\

% \section*{Acknowledgement}
\noindent\textbf{Acknowledgement} This work was supported by the Australian Research Council Grant DP200103223, FT210100228, and Australian Medical Research Future Fund MRFAI000085, Australian Future Fellowship.

{\small
\bibliographystyle{ieee_fullname}
\bibliography{egbib}
}
%--------------------------
\clearpage
\appendix

% \vspace{-8.0mm}
\section{Search Space Details}
% \vspace{-3.0mm}
\subsection{FLOPs Constraint Search Space}
\begin{table}[t]
% \centering
\begin{center}
% \small{
\begin{tabular}{>{\centering}p{2cm}|>{\centering}p{2cm}|>{\centering}p{1cm}|>{\centering}p{0.5cm}|>{\centering}p{0.5cm}}
    
    % \hline
    % \multicolumn{5}{c}{FLOPs Constrained} \tabularnewline
    % \hline
    Shape&Block&c&n&s  \tabularnewline
    \hlineB{2.5}
    
    
    $224^2 \times 3$&3x3 conv&16&1&2 \tabularnewline
    \hline
    $112^2 \times 16$&MBL&16&1&1 \tabularnewline
    $112^2 \times 16$&MBL&24&4&2 \tabularnewline
    \hline
    $56^2 \times 24$&MBL&40&4&2 \tabularnewline
%    $56^2 \times 24$&&&&&$56^2 \times 32$     && TBS    &         & 1        \\
    \hline
    $28^2 \times 40$ &MBL&80&4&2 \tabularnewline
%    $28^2 \times 40$ &&&&&$28^2 \times 64$     && TBS    &         & 1        \\
    \hline
    $14^2 \times 80$&MBL&96&4&1\tabularnewline           
    $14^2 \times 96$&MBL&192&4&2 \tabularnewline              
    \hline
    $7^2 \times 192$&MBL&320&1&1\tabularnewline   
    $7^2 \times 320$    &1x1 conv        &1024   &1  &1 \tabularnewline    
    $7^2 \times 1024$   & 7x7 avgpool &- &1  &1  \tabularnewline   
    $1024$ &fc&1024&1&- \tabularnewline   
    % \hline
\end{tabular}
        % }
\end{center}
\caption{Macro-architecture for FLOPs constraint setting. ``MBL'' denotes the learnable Multi-Branch layer, c, n, s refer to the number of backbone filters, number of layers and the stride, respectively.}
\label{tab:arch1}
\vspace{-3mm}
\end{table}

\begin{figure}[t]
\begin{center}
    \includegraphics[height=9cm]{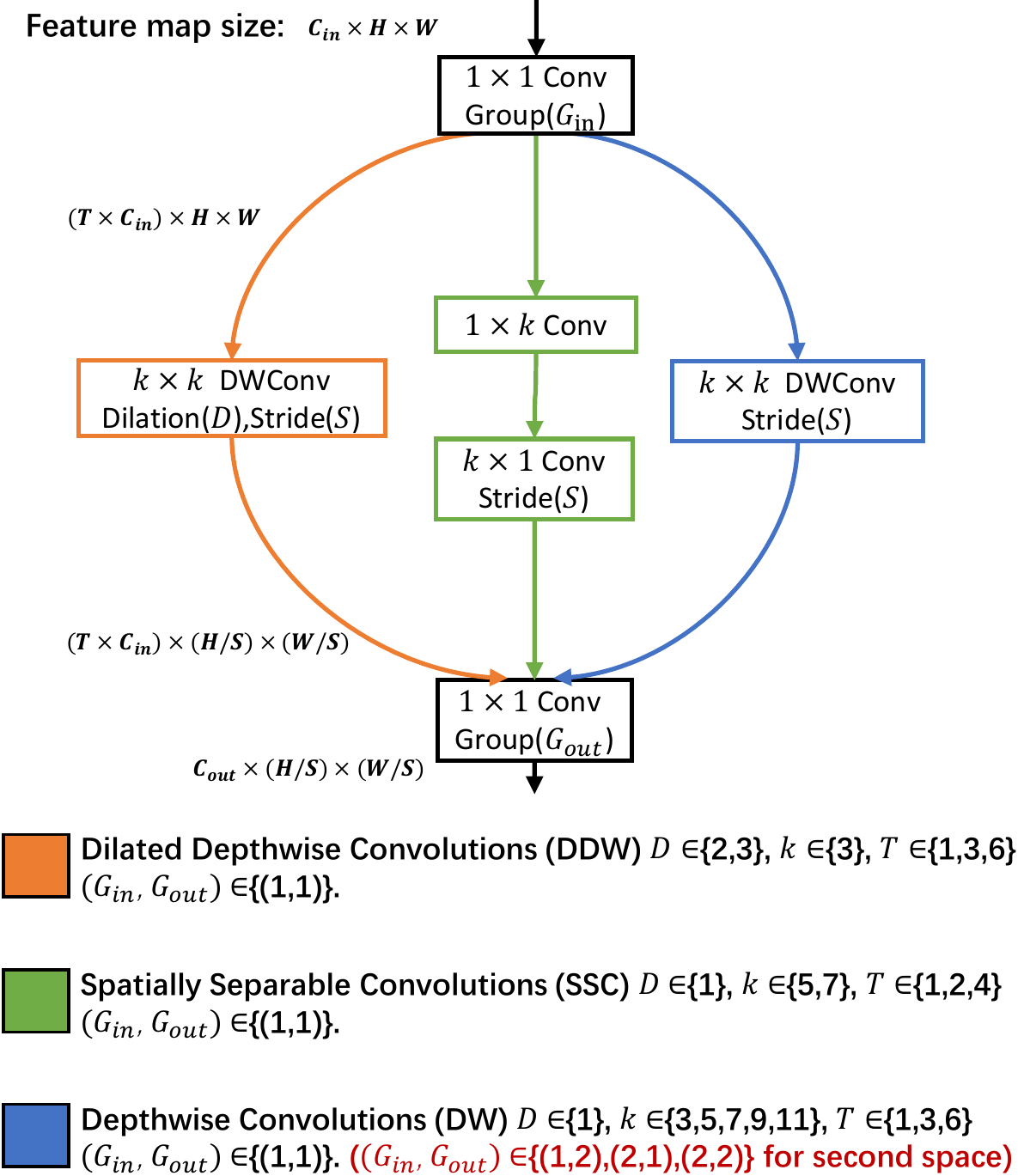}
    \caption{
    FLOPs Constraint search space details for all possible operations except Identity Mapping. The corresponding layer is recognized as reduction layer when $C_{in} \neq C_{out}$. Each type of operation has its corresponding kernel size $k$, dilation $D$, and expand ratio $T$.
    We do not search the group number $(G_{in}, G_{out})$ for $1 \times 1$ projections in the primitive 27 OPs space, for the second space, we search $G$ in  DW convolutions with varies $k$ and $T$.
    }
    \label{fig:f1}
\end{center}
\end{figure}

The 27 OPs space for FLOPs constraint, as shown in Figure~\ref{fig:f1}, is derived from multiple groups of operation designs. The first group of operations is depthwise (DW) convolution with kernel size $\{3, 5, 7, 9, 11\}$ and expand ratio $\{1, 3, 6\}$. The second group is $3 \times 3$ dilated convolution with dilation $\{2, 3\}$ and expand ratio $\{1, 3, 6\}$, this kind of operation, according to the study in MixNet \cite{tan2019mixconv}, is not efficient under FLOPs constrained scenarios. However, we still include them in our search space to test the robustness of the proposed method and see if it can find competitive architectures in a noised large search space. We also include the $1\times k-k\times1$ convolutions with $k \in \{5, 7\}$ and expand ratio $\{1, 2, 4\}$, this operation is derived from the Inception-ResNet \cite{szegedy2017inception} and is a rarely included operation in NAS literature as well.  Our major experiments are conducted in this setting. 

The second space as shown in Figure~\ref{fig:f1} consists of DW convolutions with grouped $1 \times 1$ projections, a special variant of standard DW convolution that is included in FBNet \cite{wu2018fbnet} and MixNet \cite{tan2019mixconv}. The options of kernel size and expand ratio for this variant are  $\{3, 5, 7, 9, 11\}$ and $\{1, 3, 6\}$ respectively, which is identical with standard DW convolutions in 27 OPs space.  For both search space, we use identical macro-architecture as shown in Table \ref{tab:arch1}.
% \vspace{-3.0mm}
% \vspace{-5mm}
\vspace{-3mm}
% \vspace{-8mm}
        % \vspace{-10mm}
\subsection{Latency Constraint Search Space}
\begin{figure}[t]
    \centering
    \includegraphics[height=10cm]{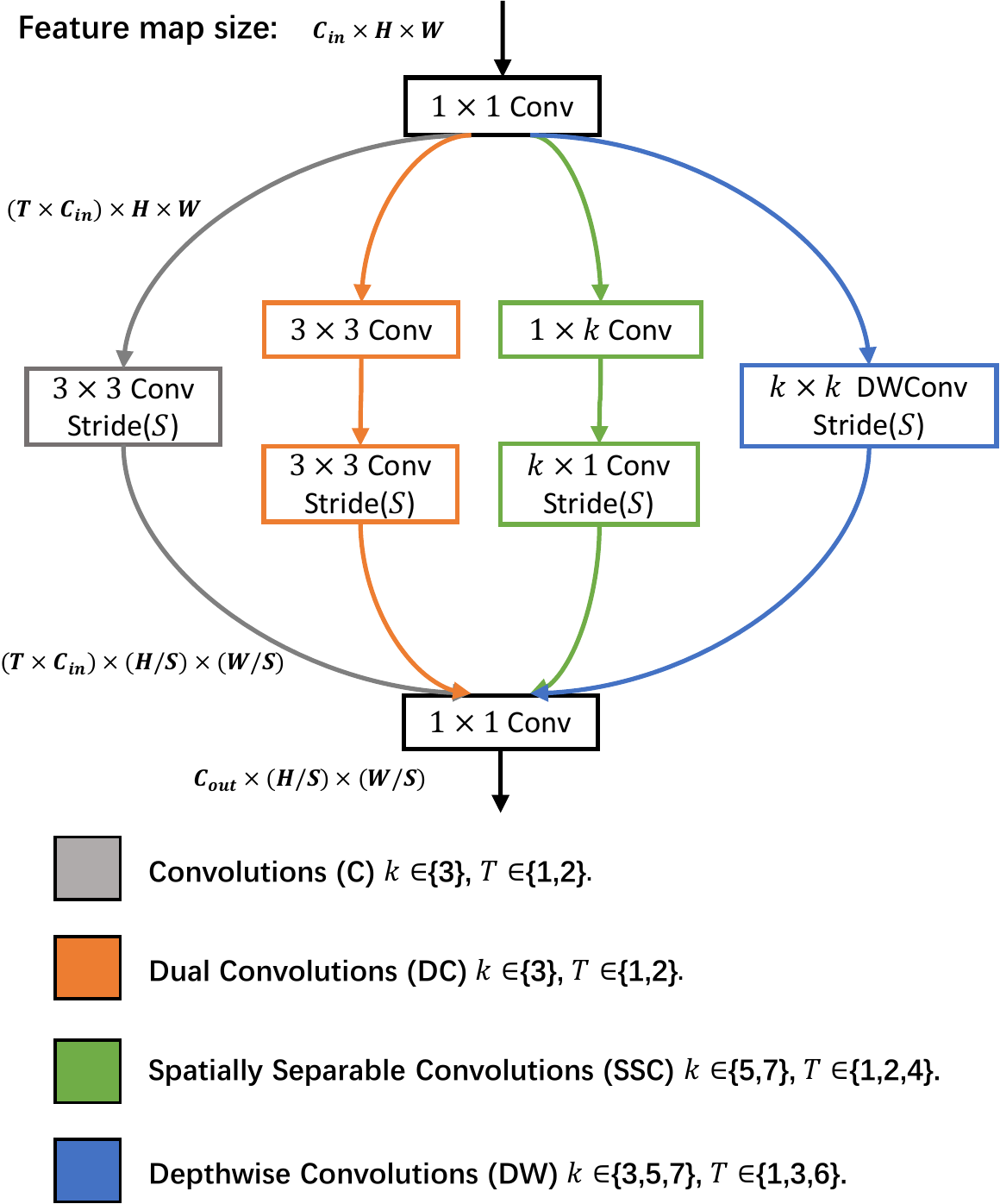}
    \caption{Latency Constraint search space details for all possible operations except Identity. These details are identical with the extended search space used by Li \etal~\cite{li2020improving}.  The corresponding layer is recognized as a reduction layer when $C_{in} \neq C_{out}$.  Each type of operation has different kernel size $k$ or expand ratio $T$. }%\wl{What is the meaning when K is a set?}
    \label{fig:f2}

\end{figure}

% \vspace{-5mm}
% \vspace{-3.0mm}
Our search space for Latency constraint as shown in Figure~\ref{fig:f2} and Table~\ref{tab:arch2} is identical with the extended search space used by Li \etal~\cite{li2020improving}.

\begin{table}[t]
\begin{center}
\begin{tabular}{>{\centering}p{2cm}|>{\centering}p{2cm}|>{\centering}p{1cm}|>{\centering}p{0.5cm}|>{\centering}p{0.5cm}}
    
    % \hline
    % \multicolumn{5}{c|}{FLOPs Constrained}&\multicolumn{5}{c}{Latency Constrained}\\
    % \hline
    Shape&Block&c&n&s \tabularnewline
    \hlineB{2.5}
    
    
    $224^2 \times 3$ &3x3 conv &16 &1 &2 \tabularnewline
    \hline
    $112^2 \times 16$ &MBL &16 &1 &1 \tabularnewline
    $112^2 \times 16$ &MBL &32 &4 &2 \tabularnewline
    \hline
    $56^2 \times 32$ &MBL &64 &4 &2 \tabularnewline
%    $56^2 \times 24$&&&&&$56^2 \times 32$     && TBS    &         & 1        \\
    \hline
    $28^2 \times 64$ &MBL &128 &8 &2 \tabularnewline
%    $28^2 \times 40$ &&&&&$28^2 \times 64$     && TBS    &         & 1        \\
    \hline
    $14^2 \times 128$ & MBL & 256 &   4 &2 \tabularnewline         
    % $14^2 \times 96$&MBL&192&4&2& & & & & \\                
    \hline
    $7^2 \times 256$  &   1x1 conv &1024 &  1 & 1\tabularnewline   
    $7^2 \times 1024$ &7x7 avgpool & - & 1 &1 \tabularnewline
    $1024$ & fc&1024 & 1 &-\tabularnewline
    % \hline
\end{tabular}
\end{center}
\caption{Macro-architecture for Latency constraint setting. ``MBL'' denotes the learnable Multi-Branch layer, c, n, s refer to the number of backbone filters, number of layers and the stride, respectively.}
% \vspace{-3mm}
\label{tab:arch2}
\end{table}

\vspace{-3mm}
\subsection{Identity Mapping Path}
% The design of our search space is partially inspired by the Inception-Resnet \cite{szegedy2017inception} and 
Inspired by the Inception-Resnet \cite{szegedy2017inception}, our search space has a residual structure, which means that all normal layers in the network have an identity mapping path (identity operation).
The identity mapping path will always be sampled during supernet training and its path probability $p$ is fixed to be $1$ during fitness indicator updates.
% \vspace{-3.8mm}
\subsection{Search Space Size Computation}
\begin{table}[t]
% \vspace{-11mm}
\begin{center}
    \begin{tabular}{>{\centering}p{2.5cm}|>{\centering}p{2cm}|>{\centering}p{2cm}}
    % \hline\noalign{\smallskip}
     & Structure &  Size\tabularnewline
    % \noalign{\smallskip}
    \hlineB{2.5}
    % \noalign{\smallskip}
    NASNet \cite{zoph2018learning}  & cell-based & $7.1 \times 10^{16}$\tabularnewline
    Amoeba \cite{real2018regularized} & cell-based &$5.6 \times 10^{14}$\tabularnewline
    ENAS \cite{pham2018efficient}& cell-based & $5.0 \times 10^{12}$\tabularnewline
    DARTS \cite{liu2018darts} & cell-based & $2.4 \times 10^{11}$\tabularnewline
    Proxyless \cite{cai2018proxylessnas} & single-branch& $3.0 \times 10^{17}$ \tabularnewline
    SPOS \cite{guo2019single} & single-branch & $1.1 \times 10^{12}$ \tabularnewline
    NSE $\ast$  & multi-branch & $1.4 \times 10^{110}$ \tabularnewline% \ast
    % \hline
    \end{tabular}
\end{center}
    \caption{ImageNet NAS search space size compared. $\ast$ when we use 27 OPs space and $K = 5$.} % \ask
    \vspace{-3mm}
\label{table:space}
\end{table}

For the case when we use 27 OPs space and layer-wise space size $K = 5$, the number of possible architectures $Comb_{arch}$ is computed as follows:

We denote the number of $k$-combinations given $n$ elements as $C^n_k = \frac{n!}{k!(n-k)!} $.     
The number of possible combinations is $Comb_{norm} = \sum_{k=0}^{5}C^{27}_k$ for the normal layer and $Comb_{redu} = \sum_{k=1}^{5}C^{27}_k$ for the reduction layer.
% For the normal layer, the number of possible combinations is $Comb_{norm} = \sum_{k=0}^{5}C^{27}_k$. As for the reduction layer, $Comb_{redu} = \sum_{k=1}^{5}C^{27}_k$, 
There are in total 16 normal layers and 6 reduction layers in FLOPs constrained macro architecture. Each layer has its own selected candidate operations. Thus the total number of possible architectures is $Comb_{arch} = (Comb_{norm})^{16} \times ( Comb_{redu})^6 \approx 1.4\times10^{110}$.
% 
% \vspace{-5mm}
\section{Details of Training Configs}

For every supernet training, we use Nesterov SGD with  0.9 momentum, weight decay $4e^{-5}$, batch size 1024 with 100 epochs. The initial learning rate is 0.4 and gradually reaches 0 through cosine learning rate decay with warm-up for 2 epochs . We use Adam optimizer with an initial learning rate of 0.1 to update fitness indicators, and we perform such updates every two supernet updates. Fitness indicators $\Theta$ are initialized to 0 and the corresponding pruning threshold is set to -2. While a larger random sample number $D$ helps to find better Pareto front, limited by its computational cost, we set sample sizes as $D=2000, D_e=100$.

For the hyperparameters of the resource constraint regularization, we set $\alpha=1e^{-5}, \beta=2, \tau=300$ for FLOPs (M) constraint and $\alpha = 2e^{-2}, \beta = 2, \tau=7$ for Latency (ms) constraint. The $\alpha$ parameter for Latency constraint is set higher so that two constraints are of similar magnitude.

For model retraining, we increase the number of epochs to 350, with batch size 2048, learning rate 0.8, weight decay $4e^{-5}$~\cite{li2020improving} for FLOPs constraint and $1e^{-4}$~\cite{liu2018progressive} for Latency constraint, together with exponential moving average with decay 0.9999. For a fair comparison, swish activation, SE module together with identical training configs from EfficientNet \cite{tan2019efficientnet} are optionally used subject to the specific settings.% as shown in Table \ref{tab:FLOPs}.

\section{Ablation on Pruning Threshold}
\begin{figure}[t]
    \centering
    \subfigure[]{
    \centering
    \includegraphics[width=0.95\linewidth]{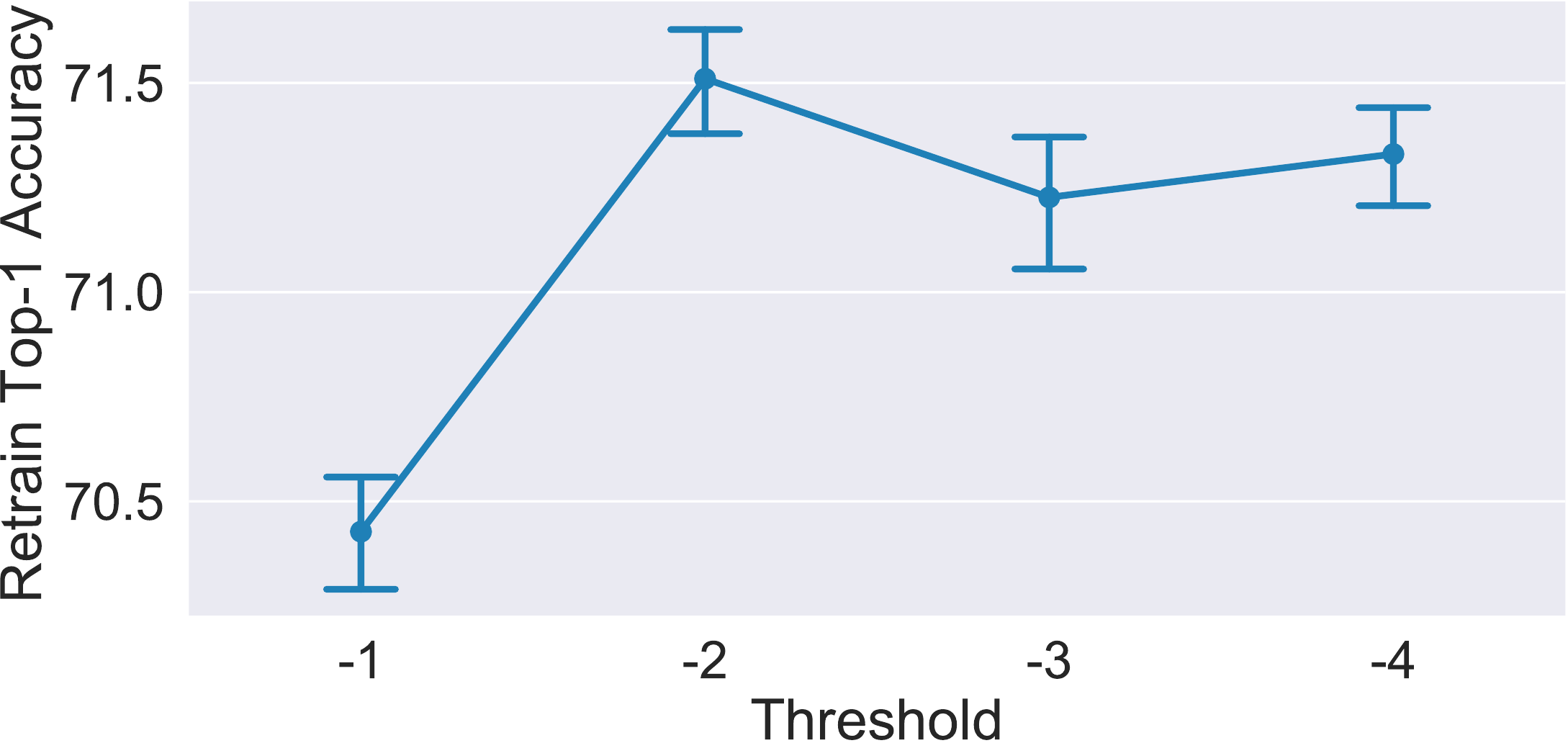}}
    % \scalebox{0.4}{\input{reduce.png}}}
    \subfigure[]{
    \centering
    \includegraphics[width=0.95\linewidth]{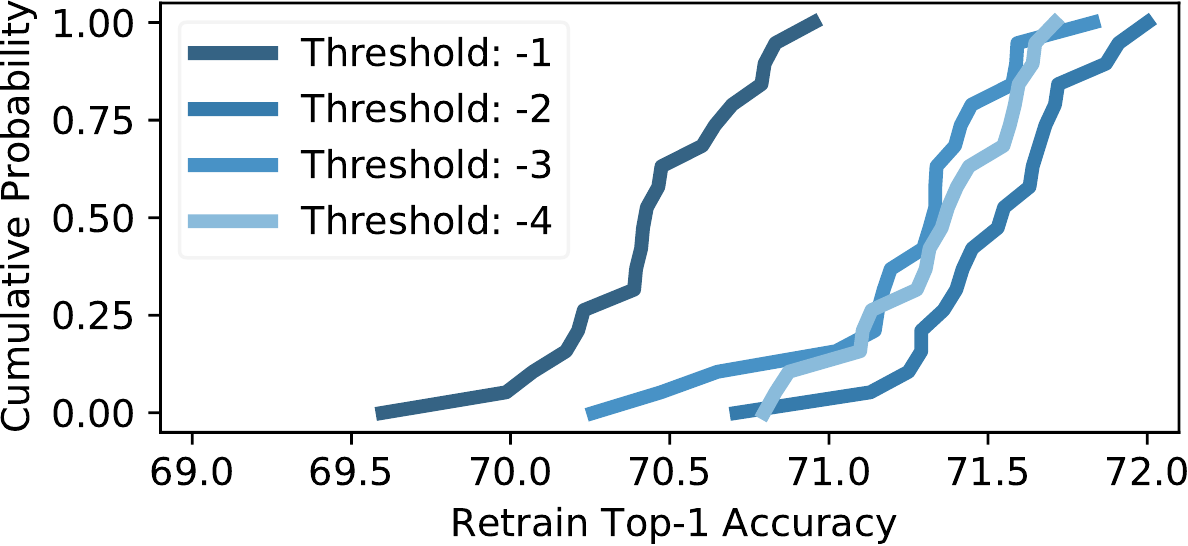}}
    % \scalebox{0.4}{\input{random.png}}}

    \caption{Comparison of 1-st round "aggregated" search space with respect to different pruning thresholds. ``aggregated" denote the search space achieved by NSE after Pareto front aggregation. All results are based on identical search space initialization with layer-wise space size $K = 5$. For each "aggregated" search space, we randomly sample 20 architectures that have FLOPs within the interval of [323M, 327M]. Each model is then trained from scratch for 50 epochs to retrieve the retrain Top-1 accuracy illustrated above. (a) accuracies are shown in mean with 95\% confidence intervals.}
    \vspace{-2mm}
        \label{fig:thres}
\end{figure}
To show how the trade-off between early and accurate search space simplification affects the optimized search space to be inherited, we evaluate the quality of aggregated search space achieved by different pruning thresholds in Figure~\ref{fig:thres}. As the threshold -1 is too close to 0 (the initialized value of fitness indicators $\Theta$), its result is significantly worse when compared to lower thresholds. However, as the threshold is set lower than -2, the result seems saturated, and a lower threshold could even harm the quality of optimized search space. 

\section{Detection Result for NSENet}

We have also evaluated our NSENet on object detection task. We take the pretrained NSENet as a drop-in replacement for the backbone feature extractor in EfficientDet-D0~\cite{tan2020efficientdet}. Table \ref{table:det} shows the performance of our NSENet, comparing with MobileNetV2 and the original backbone network EfficientNet-B0. We trained the network with identical configs as used by EfficientDet-D0. As shown in Table \ref{table:det}, our model significantly improves mAP score over MobileNetV2 and EfficientNet-B0 with fewer FLOPs. 

\begin{table}[t]
\small
% \vspace{-11mm}
\begin{center}
    \begin{tabular}{p{3cm}|>{\centering}p{2cm}|>{\centering}p{2cm}}
    % \hline%\noalign{\smallskip}
    Backbone &  FLOPs & mAP\tabularnewline
    %\noalign{\smallskip}
    \hlineB{2.5}
    %\noalign{\smallskip}
    EfficientNet-B0~\cite{tan2019efficientnet} & 2.50B &33.8~\cite{tan2020efficientdet} \tabularnewline
    MobileNetV2 1.0~\cite{sandler2018mobilenetv2} & 2.24B &32.7 \tabularnewline
    \textbf{NSENet} & \textbf{2.18B} &\textbf{34.5} \tabularnewline
    % \hline
    \end{tabular}
\end{center}
    \caption{NSENet object detection performance on COCO \cite{lin2014microsoft} dataset. All experiments adopt identical configs as used by EfficientDet-D0 \cite{tan2020efficientdet} except backbone network.} % \ask
    % \vspace{-3mm}
\label{table:det}
\end{table}

\section{Omitted Figures}
\begin{figure*}[h]
    \centering
    \includegraphics[height=5.45cm]{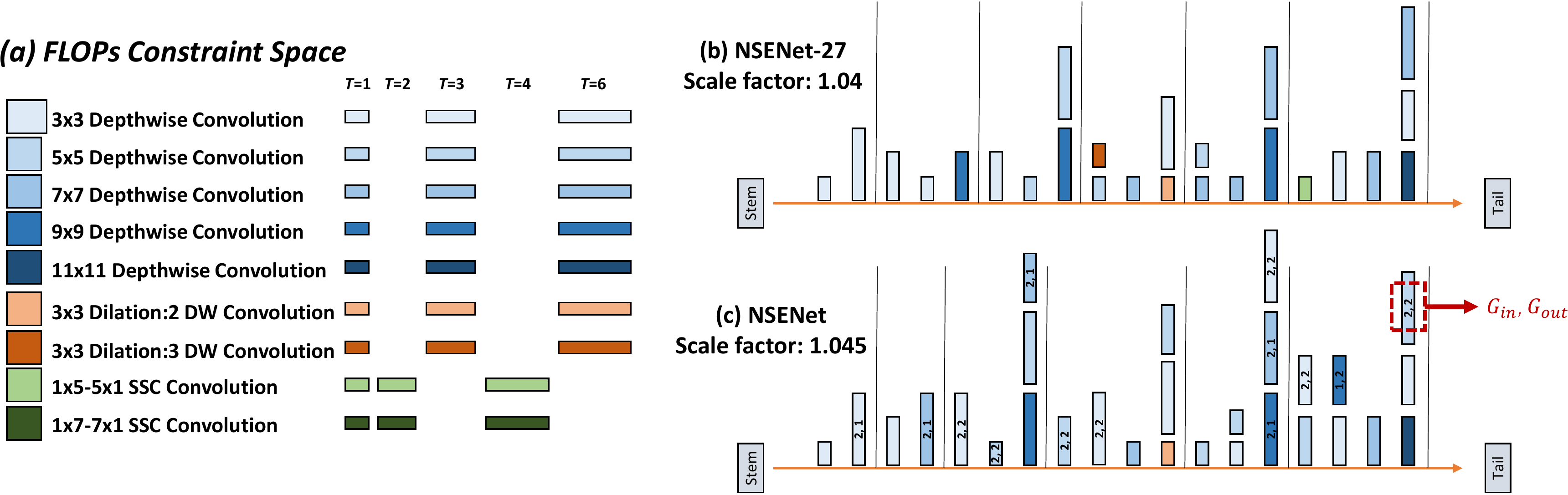}
    \vspace{5pt}
    \hrule
    \vspace{6pt}
    \includegraphics[height=3.7cm]{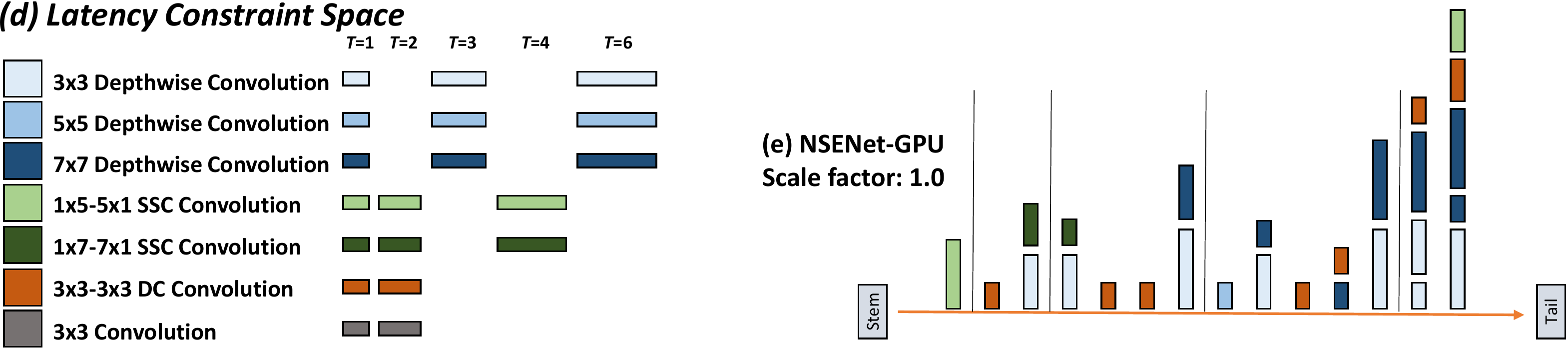}
    \caption{The detailed operations (a)(d) and structure (b)(c)(e) of our final results for FLOPs constriant and latency constraint, notice that (b) is the final result derived from 27 OPs space while (c) inherits the final search space subset derived from the 27 OPs space, then search on the second space as shown in Figure~\ref{fig:f1}. The two numbers within the operation blocks shown in (c) represents the group number $(G_{in}, G_{out})$ of 1x1 projections. The width of the blocks correspond to the $T$ in (a)(d) for candidate operation, which denotes the expand ratio of the corresponding operation, with details in Figure~\ref{fig:f1} and Figure~\ref{fig:f2}. 
    A straight line is put after every reduction layer in (b)(c) and (e). A "Scale Factor" \cite{howard2017mobilenets} is used to adjust the amount of resource (\eg FLOPs) consumed by the architecture by changing the number of channels uniformly. We can see that architectures searched under FLOPs constraint tend to go deeper while both constraints prefer efficient operations such as DW convolutions over less commonly used operations such as SSC convolutions.}
    \label{fig:nets}
\end{figure*}

\begin{figure*}[h]
    \centering
    \includegraphics[height=11.3cm]{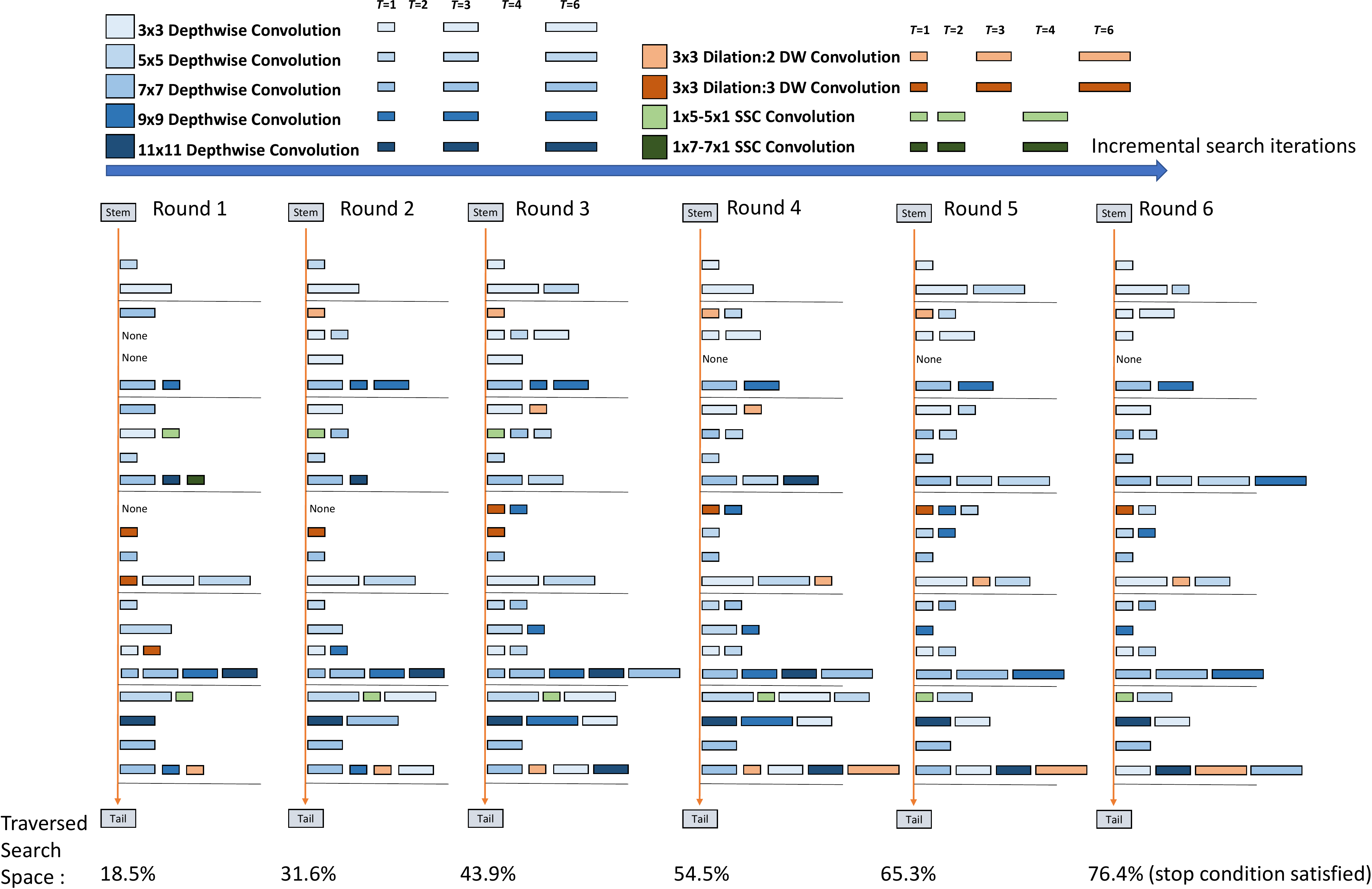}
    \caption{Intermediate results of the search space subset derived from Pareto front architecture aggregation. The results are based on the 27 OPs space and are from the same experiment where we get the NSENet-27 architecture. We can see that less commonly used operations such as SSC convolutions and dilated DW convolutions are seldom in the search space subset. On the other hand, most of the operations being included in the search space subset would last for multiple rounds or even till the final round, demonstrating the effectiveness of the proposed pipeline in terms of knowledge extraction and preservation.}
    \label{fig:int}

\end{figure*}

\begin{figure*}[h]
    \centering
%     \includegraphics[height=3.0cm]{IMG_20200228_181514.jpg}
    % \scalebox{0.8}{\input{edge.pdf_tex}}
    \includegraphics[height=10.0cm]{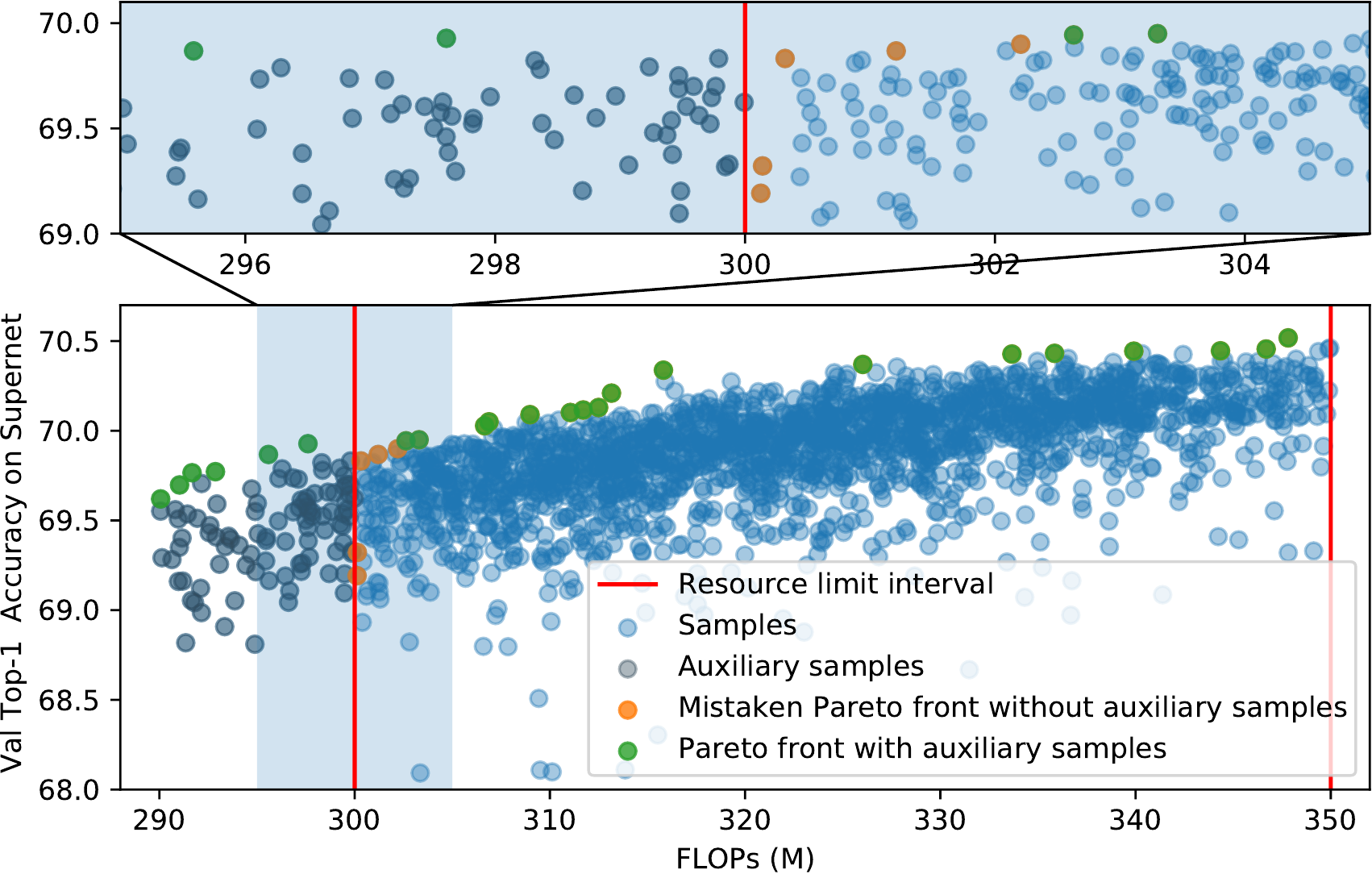}
    \caption{
    Edging effect in constrained Pareto frontier retrieval. When trying to get Pareto-optimal architectures only with the samples within the constraint interval, some of the samples (orange points in this figure) located close to the limit boundary (300M FLOPs) could be mistakenly considered as Pareto-optimal architectures. 
    By considering auxiliary samples outside the limit interval, we can alleviate this issue.
    The data used in this figure is derived from the final round of search over 27 OPs space.}
    \label{fig:edge}

\end{figure*}

Below we show the omitted figures. Figure~\ref{fig:nets} shows architecture details for final results. Figure~\ref{fig:int} shows intermediate results of aggregated search space subset on 27 OPs space. Figure~\ref{fig:edge} illustrates the edging effect on Pareto frontier.
\newpage
% \section{Architecture Details for Final Results}
% \vspace{-8.5mm}

%\wl{where are the two numbers? What is T? What is the meaning of the width of operation? Text in the figure is too small to read}

% \newpage
% ~
% \newpage
% \section{Intermediate Results of Aggregated Search Space Subset for 27 OPs Search Space}

% \pagebreak
% ~
% \pagebreak
% \section{Edging Effect in Pareto Frontier}
% \vspace{12.0mm}

\end{document}